\documentclass[sigconf]{aamas} 

\usepackage{balance} 

\usepackage{utfsym}
\usepackage{algpseudocode}
\usepackage[linesnumbered, ruled]{algorithm2e}
\usepackage{arydshln}
\usepackage{centernot}

\usepackage{epsfig}
\usepackage{subfig}
 \usepackage{pdfpages}

\setcopyright{ifaamas}
\acmConference[AAMAS '23]{Proc.\@ of the 22nd International Conference
on Autonomous Agents and Multiagent Systems (AAMAS 2023)}{May 29 -- June 2, 2023}
{London, United Kingdom}{A.~Ricci, W.~Yeoh, N.~Agmon, B.~An (eds.)}
\copyrightyear{2023}
\acmYear{2023}
\acmDOI{}
\acmPrice{}
\acmISBN{}

\acmSubmissionID{108}

\title[TiZero]{TiZero: Mastering Multi-Agent Football\\with Curriculum Learning and Self-Play} 

\author{Fanqi Lin}
\affiliation{
  \institution{Tsinghua University}
  \city{Beijing}
  \country{China}}
\email{lfq20@mails.tsinghua.edu.cn}

\author{Shiyu Huang}
\affiliation{
  \institution{4Paradigm Inc.}
  \city{Beijing}
  \country{China}}
\email{huangshiyu@4paradigm.com}

\author{Tim Pearce}
\affiliation{
  \institution{Microsoft Research}
  \city{London}
  \country{United Kingdom}}
\email{tim.pearce@microsoft.com}

\author{Wenze Chen}
\affiliation{
  \institution{Tsinghua University}
  \city{Beijing}
  \country{China}}
\email{cwz19@mails.tsinghua.edu.cn	}

\author{Wei-Wei Tu}
\affiliation{
  \institution{4Paradigm Inc.}
  \city{Beijing}
  \country{China}}
\email{tuweiwei@4paradigm.com}

\begin{abstract}

Multi-agent football poses an unsolved challenge in AI research. 
Existing work has focused on tackling simplified scenarios of the game, or else leveraging expert demonstrations. 
In this paper, we develop a multi-agent system to play the full 11 vs. 11 game mode, without demonstrations. 
This game mode contains aspects that present major challenges to modern reinforcement learning algorithms; {\bf multi-agent coordination}, {\bf long-term planning}, and {\bf non-transitivity}. 
To address these challenges, we present TiZero; a self-evolving, multi-agent system that learns {\bf from scratch}. 
TiZero introduces several innovations, including adaptive curriculum learning, a novel self-play strategy, and an objective that optimizes the policies of multiple agents jointly.
Experimentally, it outperforms previous systems by a large margin on the Google Research Football environment, increasing win rates by over $30\%$.
To demonstrate the generality of TiZero's innovations, they are assessed on several environments beyond football; Overcooked, Multi-agent Particle-Environment, Tic-Tac-Toe and Connect-Four.

\end{abstract}


\newcommand{\BibTeX}{\rm B\kern-.05em{\sc i\kern-.025em b}\kern-.08em\TeX}

\begin{document}


\pagestyle{fancy}
\fancyhead{}


\maketitle 



\section{Introduction}

Deep reinforcement learning (DRL) has achieved great success in many games, including Atari classics~\cite{mnih2015human,badia2020agent57,kapturowski2022human}, first-person-shooters~\cite{kempka2016vizdoom,huang2019combo, CSGO2022}, real-time-strategy titles~\cite{vinyals2019grandmaster,berner2019dota,ye2020towards}, board games~\cite{silver2016mastering,silver2018general} and card games~\cite{zha2021douzero,guan2022perfectdou}. Yet modern DRL systems still struggle in environments containing challenges such as  multi-agent coordination~\cite{rashid2018qmix,yu2021surprising,wen2022multi}, long-term planning~\cite{taiga2019bonus,zhang2020bebold,ecoffet2021first} and non-transitivity~\cite{balduzzi2019open,czarnecki2020real}. 
The Google Research Football environment (GFootball) contains all these challenges~\cite{googlefootball} and more. This paper presents the first system that successfully deals with all of them, learning to play the full 11 vs. 11 game mode from scratch. Experimentally, our method outperforms previous systems by a large margin with over $30\%$ higher winning rates.


\begin{table*}[t]
    \caption{Comparison of the complexity of current popular DRL benchmarks. GFootball 11 vs. 11 presents an increased level of complexity -- it combines competitive elements, stochasticisity, a large number of agents, a long game horizon and sparse rewards.}
\begin{center}
        \begin{tabular}{ccccccc}
            
            \toprule
             Citation & Game  & Competitive? & Agent number & Stochastic?  & Sparse reward? & Game length magnitude \\
            \midrule
            \cite{mnih2015human,kapturowski2022human} & Atari Games & $\usym{2717}$ & 1 & $\usym{2717}$ & Sometimes & Typically $10^2$  \\
            \cite{silver2016mastering} & Go & $\usym{2713}$ & 1 & $\usym{2717}$ & $\usym{2713}$ & $10^2$ \\
            \cite{cobbe2020leveraging} & Procgen & $\usym{2717}$ & 1 & $\usym{2713}$ & $\usym{2717}$ & $10^2$\\
            \cite{ye2020towards,gao2021learning} & Honor of Kings & $\usym{2713}$ & 5 & $\usym{2717}$ & $\usym{2717}$ & $\mathbf{10^3}$ \\
            \cite{li2021celebrating,fu2022revisiting,wen2022multi,wang2022individual} & GFootball Academy & $\usym{2717}$ & $<10$ & $\usym{2713}$ & $\usym{2713}$ & $10^2$ \\
            \midrule
            Our work & GFootball 11 vs. 11 & $\usym{2713}$ & \bf{10} & $\usym{2713}$ & $\usym{2713}$ & $\mathbf{10^3}$ \\
            \bottomrule
        \end{tabular}
    \end{center}
    \label{table:env_compare}
\end{table*}

GFootball has attracted the attention of many DRL researchers as it provides a test-bed for complex multi-agent control~\cite{li2021celebrating,fu2022revisiting,wen2022multi,huang2021tikick,wang2022individual}. As in the popular real-world sport of football/soccer, to win at GFootball agents must combine short-term control techniques with coordinated, long-term global strategies. Challenges in GFootball include multi-agent cooperation, multi-agent competition, sparse rewards, large action space, large observation space and stochastic environments -- a combination not present in other RL research environments. In GFootball, each agent needs both to cooperate with teammates, and compete against diverse and unknown opponents. 
Agents without the ball are hard to optimize as they do not obtain dense rewards, 
thus resulting in challenging multi-agent credit assignment problem~\cite{sunehag2017value,foerster2018counterfactual,zhou2020learning,feng2022multi}.
Moreover, GFootball transitions are stochastic, e.g. performing the same shooting action from the same state may sometimes result in a goal and sometimes a miss.
Table~\ref{table:env_compare} contrasts GFootball with other popoular RL environments, illustrating its complexity.

\begin{figure}[t]
\center
\includegraphics[width=0.85\linewidth]{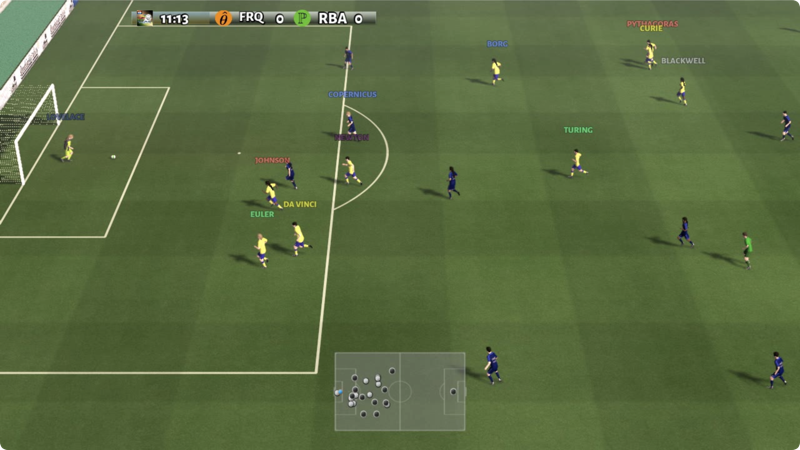}
\caption{Screenshot of Google Research Football. TiZero's agents outperform previous systems by leveraging more coordinated strategies, such as passing more often and creating more assists.}
\label{fig:football}
\end{figure}
Prior work on GFootball has mostly focused on `Academy' scenarios~\cite{li2021celebrating,fu2022revisiting,wen2022multi,wang2022individual}, which are drastically simplified versions of the full 11 vs. 11 game, for instance requiring agents to score in an empty goal or beat the goalkeeper in a 1-on-1. Any opponents are rules-based bots. 
In contrast, our work introduces an AI system that plays on the full 11 vs. 11 game mode. This mode requires simultaneous control of ten players. (The goalkeeper is excluded since they have a unique action space and different purpose which would require training of a separate policy. Instead they are controlled via a simple rule-based strategy.) Each player performs $3,000$ steps per match, meaning long-term planning is required. Due to these challenges, previous work tackling the 11 vs. 11 game mode relied on an offline demonstration dataset~\cite{huang2021tikick}. The disadvantage of such methods is that the performance of trained agents is bounded by the skill-level of the demonstrators. In this paper, we do not use any demonstration dataset nor any pre-trained model, instead training tabula-rasa via curriculum-learning and self-play.



This paper proposes a \textbf{multi-agent}, \textbf{curriculum}, \textbf{self-play} framework for GFootball, which we name TiZero, that is trained at mass-scale.
At its core, it is an actor-critic algorithm~\cite{konda1999actor}. To facilitate \textbf{multi-agent learning}, a single centralized value network guides learning of all agent policies in a joint optimization step, while allowing for decentralised execution at test time.
To tackle the challenge of long-term planning in a sparse-reward setting, we borrow ideas from \textbf{curriculum learning}~\cite{wu2016training,zheng2022local,seccia2022novel}. Here, the agent initially learns basic behaviors in simpler tasks, strengthens these in increasingly difficult scenarios, eventually enabling high-quality long-term planning. 
To address the challenge of non-transitivity (that is, player A$>$player B, player B$>$player C $\centernot\implies$ player A$>$player C)~\cite{balduzzi2019open,czarnecki2020real}, we develop a \textbf{self-play learning} strategy to manage the opponent pool, which ensures training is against a set of diverse and strong opponents. 
TiZero applies these algorithmic ideas in a large-scale distributed training infrastructure, training across hundreds of processors for about 40 days.

In addition to these core algorithmic ingredients, TiZero adopts a variety of modern DRL advances, including recurrent experience replay~\cite{kapturowski2018recurrent}, action masking, reward shaping~\cite{ng1999policy}, agent-conditioned policy~\cite{fu2022revisiting} and staggered resetting.


To evaluate the performance of TiZero, we compare against previous state-of-the-art methods~\cite{wekick,kaggle_15,kaggle_35,huang2021tikick} on GFootball, outperforming them by a large margin. 
On a public evaluation platform of GFootball, TiZero also ranks first\footnote{JiDi AI Competition Platform: \url{http://www.jidiai.cn/ranking\_list?tab=34}.\\ The evaluation result was collected on October 28th, 2022.\label{fn:public_eval}}.
To understand the generality of TiZero, we assess our system on several public benchmarks not related to football, such as ~\cite{wang2020too}, Multi-agent Particle-world (MPE)~\cite{lowe2017multi}, Tic-Tac-Toe and Connect-Four, finding that it also shows promise in these new domains.

\section{Related Work}
\label{sec:relatedwork}

In this section, we briefly review relevant work in the areas of multi-agent reinforcement learning, competitive self-play and football game AI.

\begin{figure*}[t]
\center
\includegraphics[width=0.9\linewidth]{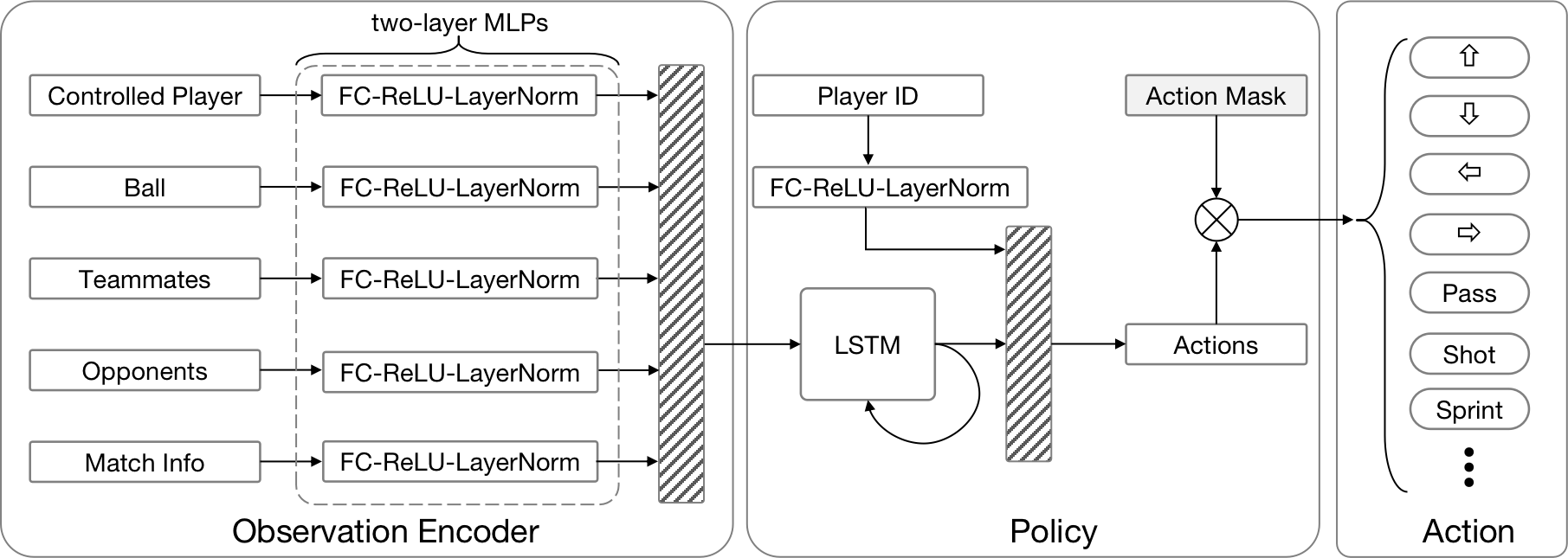}
\caption{TiZero's network architecture. Six types of information are required as input: the controlled player information, player ID, ball information, teammate information, opponent information and current match information. We use six separate MLPs with two (one for the "player ID") fully-connected layers to encode each part of the observation. An LSTM layer is used to incorporate historic observations. The policy outputs a softmax distribution over the 19 discrete actions.}
\label{fig:network}
\end{figure*}

\subsection{Multi-agent Reinforcement Learning}

Multi-agent reinforcement learning (MARL) has received much research attention~\cite{sunehag2017value,rashid2018qmix,son2019qtran,wang2020qplex,yu2021surprising}. Recent algorithms usually focus on the \emph{centralized training with decentralized execution} (CTDE) setup. In CTDE, the learning algorithm has access to the action-observation histories of all agents during training. But at execution, each agent only has access to its local action-observation history. There are two main kinds of MARL algorithms: 1) value-based algorithms, such as QMIX~\cite{rashid2018qmix}, VDN~\cite{sunehag2017value}, QPLEX~\cite{wang2020qplex}; 2) policy-based algorithms, such as MADDPG~\cite{lowe2017multi}, MAPPO~\cite{yu2021surprising}, MAT~\cite{wen2022multi}. 
As an example of a value-based MARL algorithm, QMIX uses a centralized Q-value network over the joint action-value function. It is designed in a way that constrains the joint Q-value to be monotonic with respect to each individual agent's Q-value (these are mixed in a additive but possibly non-linear function).
This allows decentralized execution as each agent can independently perform an argmax operation over their individual Q-values. 
TiZero builds upon MAPPO~\cite{yu2021surprising}, which adapts the standard single-agent PPO algorithm~\cite{schulman2017proximal} to the multi-agent setting by learning individual policies conditioned on local observations, and a centralized value function based on a global state. 
Several tricks have proven important to stabilize training, including Generalized Advantage Estimation (GAE)~\cite{schulman2015high}, observation normalization, value clipping, and orthogonal initialization. 
MAPPO is competitive with many MARL algorithms such as MADDPG~\cite{lowe2017multi}, RODE~\cite{wang2020rode}, and QMIX, both in terms of sample efficiency and wall-clock time. In this paper, we propose a new variant of MAPPO,  which can improve multi-agent coordination and reduce the video memory usage via multi-agent joint-policy optimization.

\subsection{Competitive Self-play}

Self-play has emerged as a powerful technique to obtain super-human policies in competitive environments~\cite{heinrich2015fictitious,lanctot2017unified,silver2017mastering,vinyals2019grandmaster,berner2019dota}. By playing against recent copies of itself, an agent can continuously improve its performance, avoiding any limitation that might otherwise be imposed by the skill-level of a demonstration dataset.
Policy-Space Response Oracle (PSRO) is a population learning framework for learning best-response agents to a mixture of previous agents with a meta-strategy solver ~\cite{lanctot2017unified}. Under this framework, Vinalys et al. \cite{vinyals2019grandmaster} proposed a league training framework to train robust agents for StarCraft (`prioritized fictitious self-play'). Meanwhile, OpenAI Five~\cite{berner2019dota} achieved super-human performance on Dota 2 via a simpler self-play strategy -- the current set of agents plays against the most recent set of agents with $80\%$ probability, and plays against past agents with $20\%$ probability. However, such empirical successes require huge computing resources, and the training process may become stuck due to the non-transitivity dilemma when strategic cycles exist~\cite{balduzzi2019open,czarnecki2020real}. Recent work solves this problem by increasing the diversity in the pool of opponent policies~\cite{liu2022neupl,liu2021unifying,chen2022dgpo}. This paper also introduces a novel self-play training strategy that improves the diversity and performance of opponent policies. We design a two-step self-play improvement scheme, with the first step to challenge the prior agent and second step to generalise against the entire opponent pool.

\subsection{Football Games and AI}
Football environments are valuable for AI research, as they blend several challenges together; control, strategy, cooperation, and competition. Aside from GFootball, several popular simulators have been proposed.
\emph{rSoccer}~\cite{martins2021rsoccer} and \emph{JiDi Olympics Football}~\cite{jidi_football} are simple environments -- players are represented as rigid bodies with a limited action space, either moving or pushing the ball. In contrast, GFootball provides a rich action space, adding mechanics such as slide-tackling and sprinting. 
The \emph{RoboCup Soccer Simulator}~\cite{kitano1997robocup,stone2005reinforcement,kalyanakrishnan2007half} and \emph{DeepMind MuJoCo Multi-Agent Soccer Environment}~\cite{liu2019emergent,liu2021motor} environments emphasis low-level robotic control, requiring manipulation of the joints of a player. Meanwhile, GFootball abstracts this away, allowing agents to focus on the challenge of developing high-level behaviors and strategies. 
A competition in the GFootball environment was hosted on Kaggle~\cite{kaggle} in 2020 that attracted over 1,000 teams. This required building an agent to control a single `in-focus' player at any one time, while teammates were controlled by an in-built rules-based AI.
The champion of the competition, named WeKick~\cite{wekick}, utilized imitation learning and distributed league training. 
In contrast to this setup, our system tackles the far more challenging task of controlling \textit{all} 10 outfield players on a team simultaneously, in a decentralized fashion.
The only prior work directly tackling this objective first collected a demonstration dataset by rolling out WeKick, and then utilized offline RL techniques to train agents. This was named TiKick, \cite{huang2021tikick}.
By contrast, our work trains agents via self-play without requiring demonstration data. Through this methodology, TiZero exceeds the performance of all previous systems in GFootball, including the best entry among 1,000 (WeKick), TiKick, and others.

\section{Preliminaries}
{\bf Multi-agent Reinforcement Learning}: We formalize the multi-agent reinforcement learning as a decentralized partially observable Markov decision process (Dec-POMDP)~\cite{bernstein2002complexity}. 
An $n$-agent Dec-POMDP can be represented as a tuple $(\mathcal{N},\mathcal{S},\mathcal{A},\mathcal{T},r,\mathcal{O},G,\gamma)$, where $\mathcal{S}$ is the state space, $\mathcal{A}$ is the action space, $\mathcal{O}$ is the observation space and $\gamma\in[0,1)$ is a reward discount factor. 
$\mathcal{N}\equiv\{1,...,n\}$ is a set of $n=|\mathcal{N}|$ agents.
At time step $t$, each agent $i\in \mathcal{N}$ takes an action $a^i\in \mathcal{A}$, forming a joint action $\mathbf{a}\in \mathbf{A}\equiv \mathcal{A}^n$. 
Agents receive an immediate reward $r(s,\mathbf{a})$ after taking action $\mathbf{a}$ in state $s$. The reward is shared by all agents.
$\mathcal{T}(s,\mathbf{a},s'):\mathcal{S}\times\mathbf{A}\times \mathcal{S}\mapsto [0,1]$ is the dynamics function denoting the transition probability. 
In a Dec-POMDP, an agent will receive its partially observable observation $o_t\in \mathcal{O}$ according to the observation function $G(s,i): \mathcal{S}\times\mathcal{A}\rightarrow \mathcal{O}$. 
Each agent has a policy $\pi^i(a^i_t|o^i_{1:t})$ to produce action $a^i_t$ from local historical observations $o^i_{1:t}$. 
We use $a^{-i}_t$ as the action of all complementary agents of agent $i$ and use a similar convention for policies $\pi^{-i}$. 
The agents' objective is to learn a joint policy $\mathbf{\pi}$ that 
maximizes their expected return $\mathbb{E}_{\left(s_t,\mathbf{a}_t\right)}\left[\sum_t \gamma^t r(s_t,\mathbf{a}_t)\right]$.

\section{Methodology}

This section introduces TiZero in detail. Firstly, we describe the agent's architecture and observation space. We then introduce a new multi-agent learning algorithm and self-play strategy. We further describe several important implementation details found to be helpful. Finally, we introduce the distributed training framework used to scale up our training.


\subsection{Agent Design}

\textbf{Observation space.} GFootball provides the observations in several formats, such as an RGB image of the game or a rendered mini-map. Our agents learn from only state vectors -- by default this is provided as a 115-dimensional vector, containing information such as player and ball coordinates. 
We follow previous work~\cite{huang2021tikick} which showed benefit in extending this vector with more auxiliary features (such as offside flags to mark potential offside teammates and relative positions among agents) to create a 268-dimensional vector. 
Instead of using this full vector directly as input, we split the agent observation into six parts, including the controlled player information, player ID, ball information, teammate information, opponent information and current match information. The value-network needs to approximate the value function for the whole team, thus we design a global state vector for the value-network with five parts, including ball information, information of ball holder, teammate information, opponent information and current match information. More details about our observation space can be found in the Appendix M. 



\textbf{Network architecture.} 
Six separate MLPs with two (one for the "player ID") fully-connected layers separately encode each part of the observation. These extracted hidden features are then concatenated together and processed by an LSTM layer~\cite{hochreiter1997long}, which provides the agent with memory. All hidden layers have layer normalization and ReLU non-linearities. We use the orthogonal matrix~\cite{saxe2013exact} for parameter initialization and the Adam optimizer~\cite{kingma2014adam}. To accelerate learning, we mask out any illegal actions by setting their probability of selection to zero. Figure~\ref{fig:network} shows the overall policy network architecture. We also construct a similar architecture for the value network, which is trained with Mean Square Error~(MSE) loss. Further hyperparameter details and also the value network structure are provided in the Appendix D.

\subsection{Multi-agent Algorithm}
\label{sec:muliagentalgorithm}

We now introduce the Joint-ratio Policy Optimization (JRPO), a modified version of MAPPO~\cite{yu2021surprising}. 
MAPPO is a direct extension of PPO~\cite{schulman2017proximal} to the multi-agent setting, with the value-network $V_{total}(s_t)$ centralized across all agents, learning from the global state. GAE can then be used to compute the global advantage function $A_\text{total}(s_t,\mathbf{a}_t)$ (abbreviated as $\hat{A_t}$). In MAPPO, this advantage function guides the improvement of each agent's policy independently $\pi^i_\theta(u^i_t|o^i_{1:t}), i\in \mathcal{N}$, where $\theta$ are the parameters of the policy network. Instead of this, we optimize the \textit{joint}-policy using a decentralized factorization:
\begin{eqnarray}
\begin{aligned}
\pi_\theta(\mathbf{a}_t|\mathbf{o}_{1:t}) \approx \prod^{n}_{i=1} \pi^i_\theta (a^i_t|o^i_{1:t}).
\end{aligned}
\label{eq:factorization}
\end{eqnarray}
This allows us to write the joint-policy objective as:
\begin{eqnarray}
\begin{aligned}
L^{CLIP}(\theta) = \hat{\mathbb{E}}_t\left[ \min \left( r_t(\theta) \hat{A_t}, \mathrm{clip} (r_t(\theta),1-\epsilon,1+\epsilon) \hat{A_t} \right)\right],
\end{aligned}
\label{eq:joint_ppo}
\end{eqnarray}
where $r_t(\theta) = \frac{\pi_\theta(\mathbf{a}_t|\mathbf{o}_{1:t})}{\pi_{\theta_{\mathrm{old}}}(\mathbf{a}_t|\mathbf{o}_{1:t})}=\prod^{n}_{i=1}\frac{ \pi^i_\theta (a^i_t|o^i_{1:t})}{ \pi^i_{\theta_{\mathrm{old}}}  (a^i_t|o^i_{1:t})}$, and $\hat{\mathbb{E}}_t[\cdots]$ is the expectation using empirical samples. And $\mathrm{clip}(\cdots)$ is a clipping function with clipping hyperparameter $\epsilon$.
This is in contrast to vanilla MAPPO, where $r_t(\theta)$ is computed on each agent's policy individually, $r_t(\theta, i) = \frac{ \pi^i_\theta (a^i_t|o^i_{1:t})}{ \pi^i_{\theta_{\mathrm{old}}}(a^i_t|o^i_{1:t})}$.
By using our factorization, all agent policies are optimized jointly.
Previous work has identified that this objective enjoys monotonic guarantees similar to PPO \cite{sun2022JMAPPO}.
Our later experiments evidence the empirical advantage of this joint-policy objective, compared to vanilla MAPPO. 
We believe that framing the objective jointly may encourage the agents to achieve better coordination. Additionally, we found it reduces memory usage and improves training speed.

\begin{figure*}[t]
\center
\includegraphics[width=0.9\linewidth]{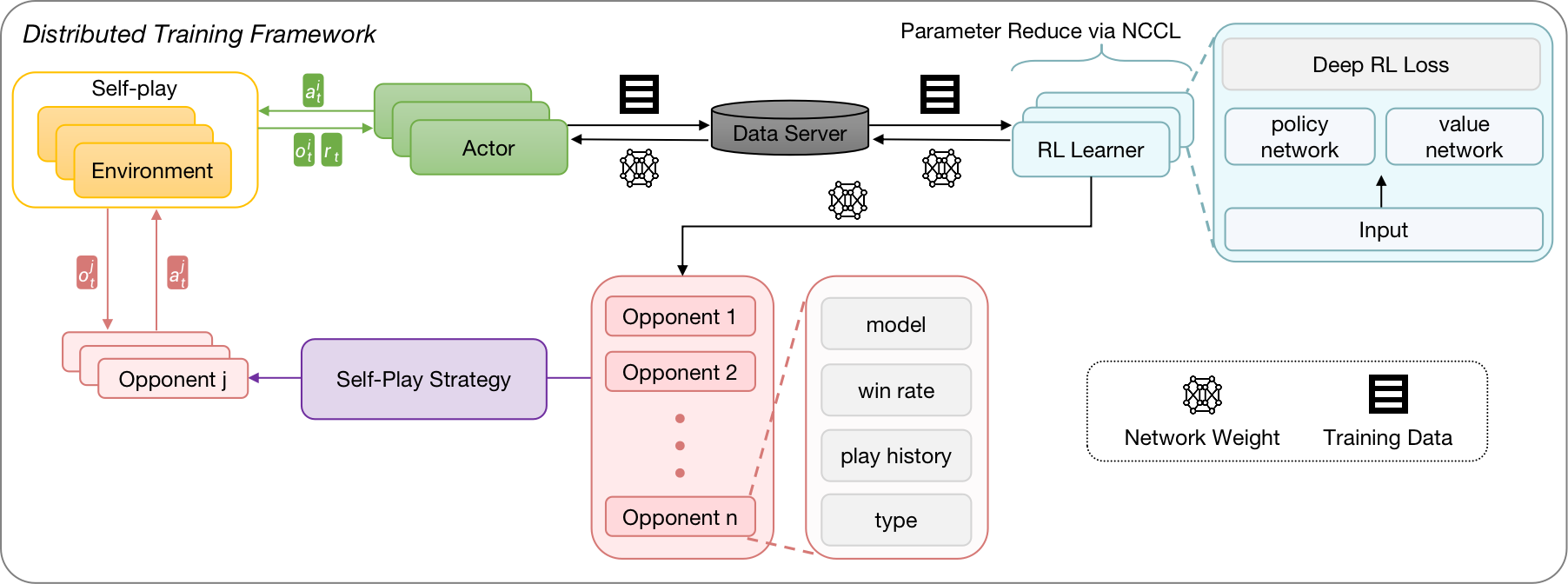}
\caption{Overview of our distributed training framework. There are two types of modules in the framework; {\bf Actors} and {\bf Learners}.
{\bf Actor} roll out models in environments, while storing observations, rewards, actions and LSTM hidden states. The data collected by the {\bf Actors} is sent to the {\bf Learners} via a data server. {\bf Learners} train the value and policy networks, and the gradients are averaged through NCCL reduction. The trained network parameters from the {\bf Learners} are synchronized to the {\bf Actors} through the data server. 
A dynamic opponent pool is maintained during training. A new opponent will be added to the pool when the training model meets certain performance criteria. A configurable self-play strategy module is used to sample opponents for self-play.}
\label{fig:distributed_framework}
\end{figure*}
\subsection{Curriculum \& Self-play Training Strategy}
\label{sec:selfplay}
TiZero's training strategy can be divided into two stages. In the first stage (`Curriculum Self-play'), the difficulty of the scenarios the agents are trained in gradually increases, guiding them to learn basic behaviors in an efficient manner. The opponents in this stage are versions of the agent trained in easier scenarios.
In the second stage (`Challenge \& Generalise Self-play'), the difficulty-level is fixed at maximum. Agents play against a diverse pool of increasingly strong opponents (previous copies of the agent), providing a route towards superhuman performance.



{\bf Curriculum Self-play}: The rewards provided in the GFootball 11 vs. 11 game mode are very sparse. This causes vanilla MARL algorithms to struggle. To address this issue, we design a curriculum self-play mechanism, in which agents are trained on a sequence of progressively more difficult scenarios, where the opponent is a copy of the agent from the previous difficulty-level scenario. We design ten difficulty levels by configuring the GFootball environment settings. The difficulty level is determined by two aspects; 1) The strength of opponent players, which can be varied from 0 to 1, with values closer to 1 meaning players are quicker and have better stamina. 2) The initial positions of players and the ball. For example, players can be set in positions closer to the opponent's goal and the ball is also set in position closer to the opponent's goal. At the beginning of training, agents are initialized with random weights, and learn on the lowest difficulty scenario (lowest opponent strength and positioned closest to the goal). This allows agents to receive denser rewards that encourage basic shooting and passing behaviors. When agent performance meets some threshold in the current scenario, the difficulty level automatically increases. Finally in the highest difficulty level, agents must compete with the whole opponent team that encourages more advanced tactics and team cooperations.
The Appendix B provides further detail about the curriculum design.

{\bf Challenge \& Generalise Self-play}:
Through curriculum self-play our agents achieve a basic performance level against a single opponent. To improve performance against a range of opponents, we design an algorithm that produces a monotonically-improving sequence of policies. This consists of two steps. 
1) {\bf Challenge Self-play.} 
Current agents play against the most recently saved agents with probability of $80\%$, and play against older versions with probability of $20\%$. The main purpose of this step is ensure the current system can defeat the strongest agents seen so far. 
2) {\bf Generalise Self-play.} 
Current agents play against the whole opponent pool, sampling opponents according to their strength as follows. Denote the opponent pool $\mathcal{M}$. Let $i \not \in \mathcal{M}$ be the current training agent, $j \in \mathcal{M}$ all other agents in pool, and $p(i,j)$ be the probability that agent $i$ defeats agent $j$. We sample model $j$ to play against with probability:
\begin{eqnarray}
\begin{aligned}
p_\text{sample}(j) = \frac{f_\text{hard}\left(p(i,j)\right)}{\sum_{m\in\mathcal{M}}f_\text{hard}\left(p(i,m)\right)},
\end{aligned}
\label{eq:sample_p}
\end{eqnarray}
where $f_\text{hard}(x)=(1-x)^2$. This sampling strategy focuses our agents training on opponents it is less likely to win against.
Therefore, the training agent will maximize its performance over all existing opponents. 
Prioritized fictitious self-play addresses the non-transitivity dilemma and improves the robustness of agents~\cite{vinyals2019grandmaster}.
Once agents perform well on this step, they are themselves added to the opponent pool for future versions to train against. 
The whole self-play algorithm and details of the our self-play strategy can be found in Appendix C \& F.

\subsection{Implementation Details}

This section summarizes several details that were found to be important to achieving good performance.


{\bf Recurrent Experience Replay}: Allowing agents to learn from earlier observations in an episode helps agents with long-term planning and inferring an opponent's strategy. Hence, we utilize an LSTM~\cite{hochreiter1997long}. Rather than initializing hidden states with zeros during training on each sequence, we reset to the hidden state that was actually generated by the agent at that timestep -- these are stored in the replay buffer during roll outs. When collecting the training data, models are rolled out for sequences of 500 timesteps. During optimization, the LSTM backpropagates through timesteps with length of 25.

{\bf Action Masking}: Action masking reduces the effective action space by preventing selection of inappropriate actions. In GFootball, we mask out actions that are unavailable or nonsensical during certain situations, for instance slide-tackling is disabled when a teammate holds the ball.


{\bf Reward Shaping}: The basic rewards provided by GFootball are $1$ for scoring a goal, and $-1$ for conceding a goal. This is sparse and hard to optimize. To improve training efficiency, we design several more dense reward signals. Note all agents on a team receive rewards equally.
\begin{itemize}
\item[-] {\bf Holding-Ball reward}: When the ball is controlled by an agents' team, a reward of $0.0001$ per timestep is received.
\item[-] {\bf Passing-Ball reward}: When agents execute successful pass before a goal, they receive a reward of $0.05$. This encourages agents to learn coordinated passing strategies.
\item[-] {\bf Grouping penalty}: If agents on the same team gather too closely together, a reward of $-0.001$ is received. This encourages agents to spread out across the pitch.
\item[-] {\bf Out-of-bounds penalty}: A reward of $-0.001$ is received when an agent is outside of the playing area. 
\end{itemize}
The reward across the two teams is balanced to be zero-sum, which is a basic requirement for self-play to succeed. Hence, if one team receives a positive reward, the other team receives a negative reward of the same magnitude. 



{\bf Player-ID Conditioning}: Instead of training a separate policy network per agent, we use a single shared policy network which receives the player ID as input. Hence, one set of parameters is used by all the agents on a team, while still allowing for decentralized execution. This makes the multi-agent training more sample efficient, since experience from one agent helps improve the network of other agents. 
It also improves inference speed -- only one forward pass of the observation encoder and LSTM is required for the whole team. This produces the embedding that is used by \textit{all} agents after appending their player-ID embedding to the output of the LSTM unit (Figure~\ref{fig:network}).





{\bf Staggered Resetting}: To encourage episodes to be initialized in interesting situations, we use the `staggered resetting' trick when collecting the training data. After the first time resetting the environment, we apply random actions for all players for a variable number of timesteps. This avoids each episode beginning from the same state, which could introduce undesirable correlations in our data.



\begin{table*}[t]
    \caption{Comparison of TiZero with baseline systems on GFootball. Higher is better except for all metrics except 'Draw Rate' and 'Lose Rate'. As well as a better win rate and goal difference, TiZero agents demonstrate a higher level of teamwork, passing the ball more and creating more assists.}
\begin{center}
        \begin{tabular}{lccccccc}
            \toprule
             Metric & TiZero (Ours)  & TiKick & WeKick & JiDi\_3rd  & Built-in Hard & Rule-Based-1 & Rule-Based-2 \\
            \midrule
            Assist & \textbf{1.30\footnotesize(1.02)} & 0.61\footnotesize(0.79) & 0.20\footnotesize(0.47) & 0.35\footnotesize(0.62) & 0.20\footnotesize(0.55) & 0.28\footnotesize(0.59) & 0.22\footnotesize(0.53)\\
            Pass & \textbf{19.2\footnotesize(3.44)} & 6.99\footnotesize(2.71) & 5.33\footnotesize(2.44) & 3.96\footnotesize(2.33) & 11.5\footnotesize(4.63) & 7.28\footnotesize(2.77) & 7.50\footnotesize(3.12)\\
            Pass Rate & \textbf{0.73\footnotesize(0.07)} & 0.65\footnotesize(0.17) & 0.53\footnotesize(0.18) & 0.44\footnotesize(0.19) & 0.66\footnotesize(0.12) & 0.64\footnotesize(0.17) & 0.63\footnotesize(0.19)\\
            Goal & \textbf{3.42\footnotesize(1.69)} & 1.79\footnotesize(1.41) & 0.88\footnotesize(0.88) & 1.43\footnotesize(1.34) & 0.52\footnotesize(0.91) & 0.73\footnotesize(0.69) & 0.64\footnotesize(0.82)\\
            Goal Difference & \textbf{2.27\footnotesize(1.93)} & 0.71\footnotesize(2.08) & -0.47\footnotesize(1.68) & -0.02\footnotesize(2.14) & -1.06\footnotesize(1.93) & -0.60\footnotesize(1.03) & -0.71\footnotesize(1.45)\\
            Draw Rate (\%) & \textbf{8.50} & 22.2 & 29.0 & 23.2 & 24.8 & 28.7 & 27.8 \\
            Lose Rate (\%) & \textbf{6.50} & 23.5 & 44.2 & 33.8 & 59.6 & 48.2 & 49.5\\
            Win Rate (\%) & \textbf{85.0} & 54.3 & 26.8 & 43.0 & 15.6 & 23.1 & 22.7\\
            \hdashline
            TrueSkill & \textbf{45.2} & 37.2 & 30.9 & 35.0 & 24.9 & 28.2 & 27.1\\
            \bottomrule
        \end{tabular}
    \end{center}
    \label{table:baseline_compare}
\end{table*}

\subsection{Distributed Training Framework}

Training a deep neural network at the scale demanded by GFootball requires a large amount of computation. In this work, we built a scalable and loosely-decoupled distributed infrastructure for multi-agent self-play training . 
There are two types of modules in our framework; {\bf Actors} and {\bf Learners}. These modules are decoupled, enabling researchers to develop their algorithm on a single local machine and then easily launch a large-scale training with {\bf zero-code change}.
The main function of the {\bf Actor} is rolling out models in environments (or simulators), while storing observations, rewards, actions and LSTM hidden states. All the data collected by the {\bf Actors} is sent to the {\bf Learners} via a data server. Learners will train the value and policy networks using GPUs and the gradients are averaged through NCCL reduction~\cite{nccl}. The trained network parameters from the {\bf Learners} are synchronized to the {\bf Actors} through the data server. 
Last but not least, we maintain a dynamic opponent pool and continuously add new opponent to the pool when the training model meets certain performance indexes. Researchers can utilize the stored opponent information (such as winning rates and historical play information) and are flexible to design their own self-play strategies to sample opponents for the self-play.
Besides, our model definition and training is done using Pytorch~\cite{paszke2019pytorch}.
Figure~\ref{fig:distributed_framework} summarises our distributed training framework.


\begin{figure}[h]
\center
\includegraphics[width=0.95\linewidth]{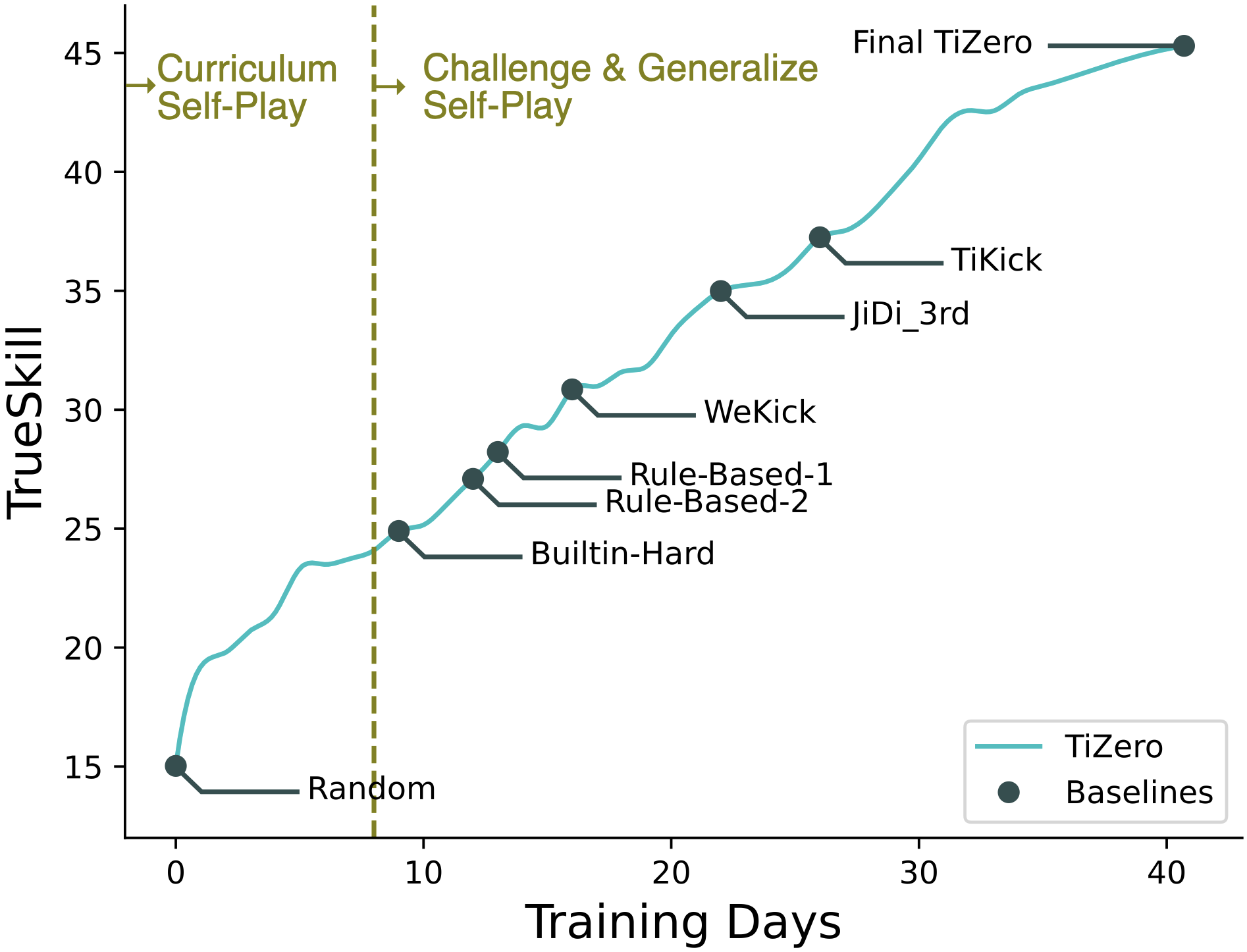}
\caption{Evolution of TiZero strength over 45 days of training. Dashed green line denotes transition from Curriculum Self-Play to Challenge \& Generalize Self-Play. We can observe that the Curriculum Self-play stage allows TiZero to rapidly improve early in training to a level just below the `Built-in Hard' agent. Through Challenge \& Generalize self-play, TiZero continues to gradually improve performance beyond the level of previous systems, eventually plateauing at a rating around 45.}
\label{fig:trueskill}
\end{figure}
\section{Experiments}

This section empirically evaluates the performance of TiZero. We first test the full system on our target domain, GFootball. We then explore the generality of several components of the system on environments unrelated to football. Specifically we show that JRPO improves over MAPPO, and that our Challenge \& Generalise Self-Play framework produces stronger and more diverse strategies than alternatives.


\begin{figure*}[t]
\center
\subfloat[]{\begin{centering}
\includegraphics[width=0.99\linewidth]{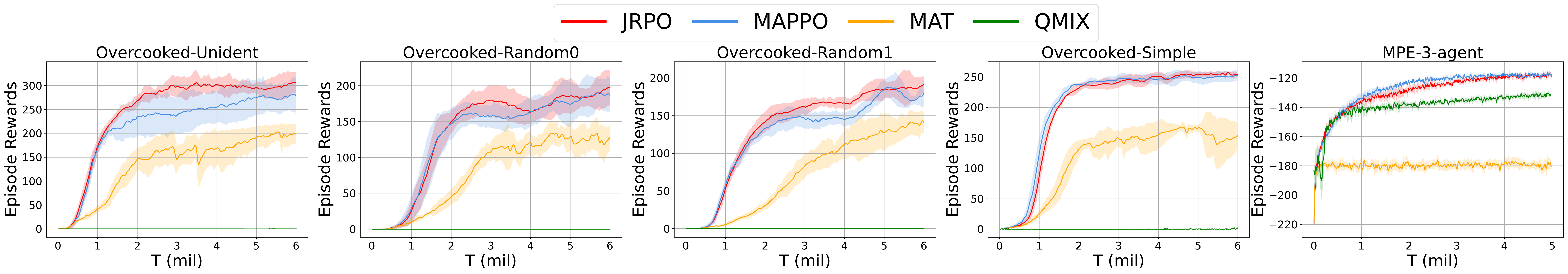}
\end{centering}
}

\subfloat[]{\begin{centering}
\includegraphics[width=0.99\linewidth]{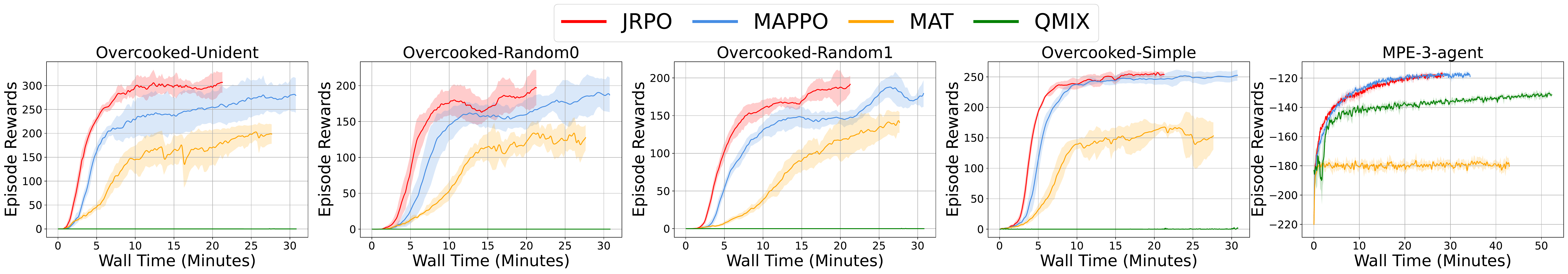}
\end{centering}
}

\caption{Comparison of multi-agent algorithms. 
(a) Training curves over environment steps. JRPO's performance is equal or better than MAPPO, MAT and QMIX across all tasks. (b) Training curves over wall-clock time. JRPO trains faster, leading to improvements in terms of wall-clock training time.}
\label{fig:marl_compare}
\end{figure*}
\subsection{GFootball Evaluation}

{\bf Experimental Settings}: 
We train and evaluate TiZero on the full 11 vs. 11 game mode in GFootball. Each match lasts for five minutes, or $3,000$ timesteps (no injury time). The team with highest number of goals wins. Standard football rules are applied by the game, such as offside, penalty kicks and yellow/red cards.
TiZero was trained over 45 days on a cluster with 800 CPUs and two NVIDIA A100 GPUs. The batch size for each GPU is set to $2,150,000$, the hidden size of the LSTM layer is 256, and the discount factor $\gamma$ is $0.999$. We used the Adam optimizer with learning rate of $0.0001$. Further hyperparameters can be found in the Appendix H. 

We compare TiZero to several strong baselines:
\begin{itemize}
    \item[-] {\bf WeKick}~\cite{wekick}: An RL-based agent that placed first from over 1,000 entries in the 2020 Kaggle Football Competition~\cite{kaggle} (see Section \ref{sec:relatedwork}). 
    \item[-] {\bf TiKick}~\cite{huang2021tikick}: The current strongest agent for the GFootball full game, controls all players in the game (see Section \ref{sec:relatedwork}). 
    \item[-] {\bf JiDi\_3rd}: An agent initialized with a pre-trained model and improved via self-play. This was the second runner-up of the 2022 JiDi Football Competition\footnote{\url{http://www.jidiai.cn/compete\_detail?compete=16}}.
    \item[-] {\bf Built-in Hard}: Agent provided by the GFootball environment, which directly controls the underlying game engine.
    \item[-] {\bf Rule-based}: Two strong rules-based agents from the Kaggle Football Competition, which were hand designed. We use Rule-Based-1~\cite{kaggle_15} and Rule-Based-2~\cite{kaggle_35}.
\end{itemize}

We use the TrueSkill rating system~\cite{herbrich2006trueskill} to evaluate all systems. 
Ratings are computed by running a large amount of matches over a fixed pool of all systems -- this pool comprises all the models checkpointed during TiZero's training process, as well as baseline systems. Figure~\ref{fig:trueskill} shows the evolution of the TrueSkill rating over 45 days of training. We also mark the TrueSkill rating of the baseline systems, all of which TiZero exceeds by a wide margin. 
One can observe that the curriculum self-play stage allows TiZero to rapidly improve early in training to a level just below the `Built-in Hard' agent. Through challenge \& generalize self-play, TiZero continues to gradually improve performance beyond the level of previous systems, eventually plateauing at a rating around 45. Table~\ref{table:baseline_compare} presents TrueSkill ratings and win rates for all systems.
To verify our system independently, we also submitted our best TiZero system to a public evaluation platform, which maintains a public leaderboard of GFootball systems. At present, TiZero ranks first with a score of $9.7$ and win rate of $95.8\%$.

We conducted an analysis to understand how cooperative TiZero's decentralized agents are. It's possible that superior performance could be achieved through expert control of a single player, rather than through coordinated teamwork leveraging strategies such as `tiki-taka', crosses and long balls.
Table~\ref{table:baseline_compare} reports statistics that reveal TiZero's agents indeed leverage cooperation more than other systems. TiZero has a higher number of passes or long balls that directly lead to scoring (`Assists'), more passes between teammates (`Pass'), and a higher chance of passes being successfully received by teammates (`Pass Rate').
The Appendix A provides detailed visualizations of TiZero's coordination behaviors.


\subsection{Policy Optimization Ablation}

This section compares the joint-policy version of PPO (JRPO) we introduced in Section \ref{sec:muliagentalgorithm}, with established state-of-the-art MARL algorithms MAPPO~\cite{yu2021surprising}, MAT~\cite{wen2022multi} and QMIX~\cite{rashid2018qmix}. 
We test across the environments Overcooked~\cite{wang2020too} and Multi-agent Particle-Environment (MPE)~\cite{lowe2017multi}. Overcooked is a grid-world game in which agents cooperate to complete a series of tasks, such as finding vegetables, making soup and delivering food. MPE is a continuous 2D world with multiple movable particles. More information about these environments can be found in Appendix G.
To be comparable, all three MARL algorithms use the same desktop machine, neural network architecture and hyperparameters. Each method is run over 6 random seeds. More details about the setup of each environment can found in the Appendix I.


\begin{table}[h]
\caption{GPU memory consumption of each method. Lower is better. Results show that JRPO consumes less GPU memory than MAPPO, especially when there are more agents (such as MPE with 20 agents and GFootball with 10 agents).}
\begin{center}
        \begin{tabular}{ccc}
            \toprule
             Task & JRPO & MAPPO\\
            \midrule
            MPE-3-agent & {\bf 1.21}GB & 1.24GB \\
            MPE-20-agent & {\bf 1.63}GB & 2.14GB \\
            Overcooked & {\bf 5.27}GB & 6.65GB \\
            GFootball & {\bf 121}GB & 196GB \\
            \bottomrule
        \end{tabular}
    \end{center}
    
    \label{table:GPU}
\end{table}

Figure~\ref{fig:marl_compare} shows the training curves of each methods w.r.t. environment steps and wall-clock time. JRPO's performance is equal or better than MAPPO, MAT and QMIX across all tasks. JRPO also achieves better results with faster wall-clock training time. 
Table~\ref{table:GPU} shows the GPU memory consumption of JRPO and MAPPO. Results show that JRPO consumes less GPU memory than MAPPO, especially when there are more agents (such as MPE with 20 agents and GFootball with 10 agents).
More experimental results can be found in the Appendix K. 



\subsection{Self-Play Strategy Ablation}

\begin{figure}[h]
\center
\includegraphics[width=\linewidth]{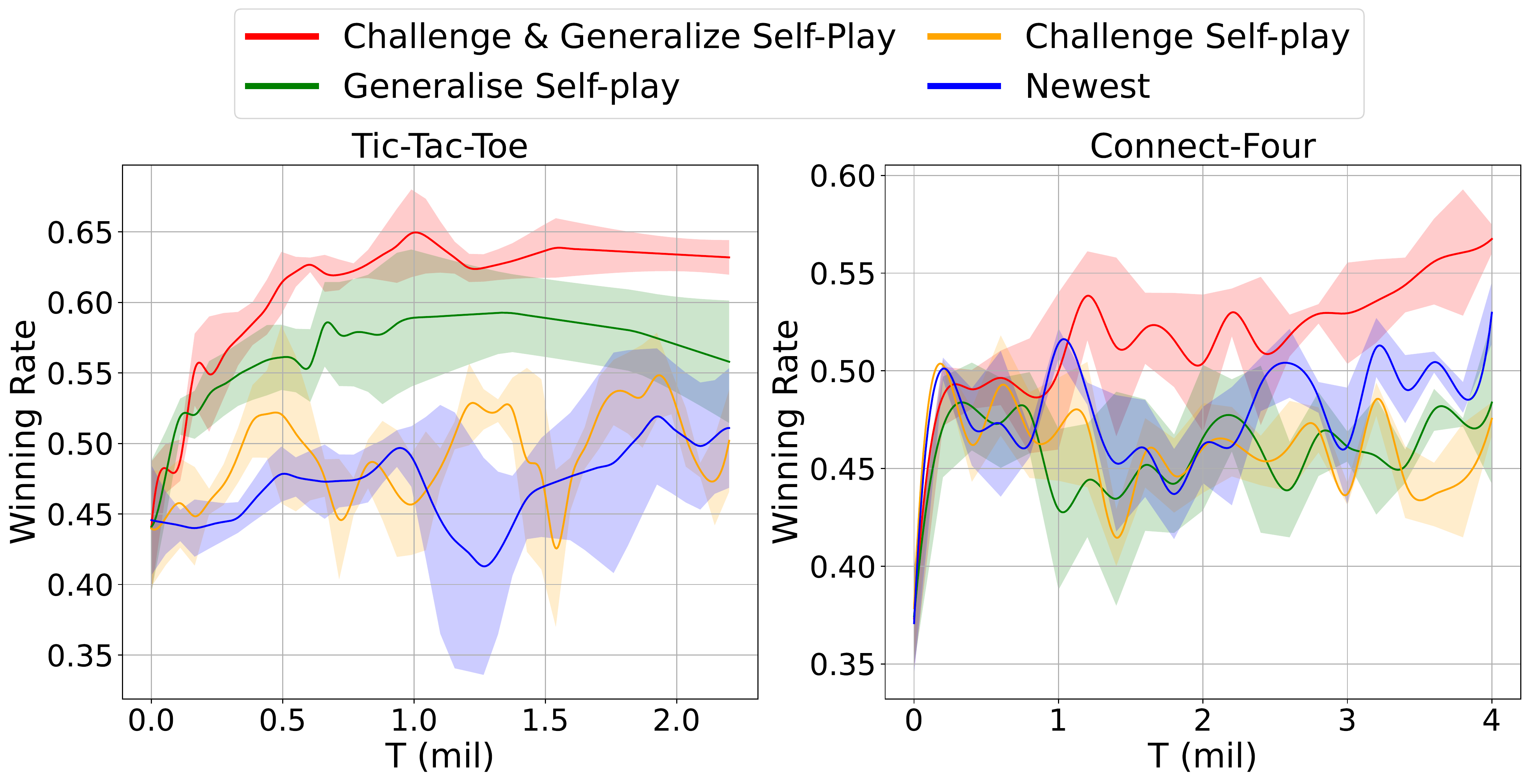}


\caption{Self-play strategy evaluation. Results show that our method outperforms other baselines in both training efficiency and final performance. It also appears more stable.}
\label{fig:seflplay}
\end{figure}

This section evaluates our `Challenge \& Generalize Self-Play' strategy on two classical 2-player board games, Tic-Tac-Toe and Connect-Four. In the Tic-Tac-Toe game, two players take turns placing marks on a three-by-three grid. Players win by placing three of their marks in a line. Our agent is a deep neural network with a 27-dimension state vector as input (O's, X's and blanks are one-hot encoded for the 9 cells). Connect-Four is an extended version of Tic-Tac-Toe with a 
four-by-four board and with placing four marks in a row to win. Our self-play strategy was described in Section~\ref{sec:selfplay}. This ablation omits the curriculum self-play stage since the rewards are non-sparse. Thus, we focus our test on the challenge \& generalize self-play stage. We construct several baselines as a comparison: (1) {\bf Challenge Self-Play}: The training agent uses the Challenge Self-Play as described in Section~\ref{sec:selfplay}; (2) {\bf Generalize Self-Play}: The training agent uses the Generalize Self-Play as described in Section~\ref{sec:selfplay}; (3) {\bf Newest}: The training agent only combats with itself. All methods share the same network architecture and learning paradigm. The only difference is how they sample opponents from the pool for self-play. 

\begin{table}[h]
    \caption{Diversity Index of policies in the opponent pool of each method, measured on Tic-Tac-Toe. Larger values indicate a higher variance of strategies in the opponent pool.}
\begin{center}
        \begin{tabular}{cc}
            \toprule
            Self-Play Sampling Strategy & Diversity Index \\
            \midrule
            Newest & 6.65\\
            Challenge Self-Play & 7.19\\
            Generalize Self-Play & 7.15 \\
            Challenge \& Generalize Self-Play & {\bf 8.11}\\
            \bottomrule
        \end{tabular}
    \end{center}

    \label{table:diversity_compare}
\end{table}

Figure~\ref{fig:seflplay} shows the training curves of different self-play strategies over three random seeds. Results show that our method outperforms other baselines in both training efficiency and final performance. 
For Tic-Tac-Toe, we quantitatively evaluate the Diversity Index\cite{parker2020effective} of policies in the opponent pool of each method, as shown in Table~\ref{table:diversity_compare}. We see that 'Challenge \& Generalize Self-Play' achieves a larger Diversity Index than baselines, which indicates our self-play strategy can produce more diverse policies.
Further experimental details and visualizations are given in Appendix J \& L.

\section{Conclusion}

This paper presented a distributed multi-agent reinforcement learning system for the complex Google Research Football environment.
This environment encapsulates several challenges;  multi-agent coordination, sparse rewards, and non-transitivity. 
Our work is the first to train strong agents for the GFootball 11 vs. 11 game mode from scratch, controlling all 10 outfield players in a decentralized fashion.
Achieving this required combining existing techniques, with several innovations -- a joint-policy optimization objective, curriculum self-play and a challenge \& generalize self-play strategy. 
Our experiments in GFootball showed that TiZero outperforms previous systems by a wide margin in terms of win rate and goal difference. TiZero also utilizes complex coordination behaviors more often than prior systems. 
Experiments on other MARL and self-play benchmarks further evidence the effectiveness and generality of our algorithmic innovations.

\newpage
\bibliographystyle{ACM-Reference-Format} 
\bibliography{sample}


\begin{thebibliography}{71}


\ifx \showCODEN    \undefined \def \showCODEN     #1{\unskip}     \fi
\ifx \showDOI      \undefined \def \showDOI       #1{#1}\fi
\ifx \showISBNx    \undefined \def \showISBNx     #1{\unskip}     \fi
\ifx \showISBNxiii \undefined \def \showISBNxiii  #1{\unskip}     \fi
\ifx \showISSN     \undefined \def \showISSN      #1{\unskip}     \fi
\ifx \showLCCN     \undefined \def \showLCCN      #1{\unskip}     \fi
\ifx \shownote     \undefined \def \shownote      #1{#1}          \fi
\ifx \showarticletitle \undefined \def \showarticletitle #1{#1}   \fi
\ifx \showURL      \undefined \def \showURL       {\relax}        \fi
\providecommand\bibfield[2]{#2}
\providecommand\bibinfo[2]{#2}
\providecommand\natexlab[1]{#1}
\providecommand\showeprint[2][]{arXiv:#2}

\bibitem[\protect\citeauthoryear{Anvarov}{Anvarov}{2020}]%
        {kaggle_35}
\bibfield{author}{\bibinfo{person}{Sarvar Anvarov}.}
  \bibinfo{year}{2020}\natexlab{}.
\newblock \bibinfo{title}{Solution ranked 35th in Kaggle Football Competition}.
\newblock
  \bibinfo{howpublished}{\url{https://github.com/Sarvar-Anvarov/Google-Research-Football}}.
\newblock


\bibitem[\protect\citeauthoryear{Badia, Piot, Kapturowski, Sprechmann,
  Vitvitskyi, Guo, and Blundell}{Badia et~al\mbox{.}}{2020}]%
        {badia2020agent57}
\bibfield{author}{\bibinfo{person}{Adri{\`a}~Puigdom{\`e}nech Badia},
  \bibinfo{person}{Bilal Piot}, \bibinfo{person}{Steven Kapturowski},
  \bibinfo{person}{Pablo Sprechmann}, \bibinfo{person}{Alex Vitvitskyi},
  \bibinfo{person}{Zhaohan~Daniel Guo}, {and} \bibinfo{person}{Charles
  Blundell}.} \bibinfo{year}{2020}\natexlab{}.
\newblock \showarticletitle{Agent57: Outperforming the atari human benchmark}.
  In \bibinfo{booktitle}{\emph{International Conference on Machine Learning}}.
  PMLR, \bibinfo{pages}{507--517}.
\newblock


\bibitem[\protect\citeauthoryear{Balduzzi, Garnelo, Bachrach, Czarnecki,
  Perolat, Jaderberg, and Graepel}{Balduzzi et~al\mbox{.}}{2019}]%
        {balduzzi2019open}
\bibfield{author}{\bibinfo{person}{David Balduzzi}, \bibinfo{person}{Marta
  Garnelo}, \bibinfo{person}{Yoram Bachrach}, \bibinfo{person}{Wojciech
  Czarnecki}, \bibinfo{person}{Julien Perolat}, \bibinfo{person}{Max
  Jaderberg}, {and} \bibinfo{person}{Thore Graepel}.}
  \bibinfo{year}{2019}\natexlab{}.
\newblock \showarticletitle{Open-ended learning in symmetric zero-sum games}.
  In \bibinfo{booktitle}{\emph{International Conference on Machine Learning}}.
  PMLR, \bibinfo{pages}{434--443}.
\newblock


\bibitem[\protect\citeauthoryear{Berner, Brockman, Chan, Cheung, Debiak,
  Dennison, Farhi, Fischer, Hashme, Hesse, et~al\mbox{.}}{Berner
  et~al\mbox{.}}{2019}]%
        {berner2019dota}
\bibfield{author}{\bibinfo{person}{Christopher Berner}, \bibinfo{person}{Greg
  Brockman}, \bibinfo{person}{Brooke Chan}, \bibinfo{person}{Vicki Cheung},
  \bibinfo{person}{Przemyslaw Debiak}, \bibinfo{person}{Christy Dennison},
  \bibinfo{person}{David Farhi}, \bibinfo{person}{Quirin Fischer},
  \bibinfo{person}{Shariq Hashme}, \bibinfo{person}{Chris Hesse},
  {et~al\mbox{.}}} \bibinfo{year}{2019}\natexlab{}.
\newblock \showarticletitle{Dota 2 with large scale deep reinforcement
  learning}.
\newblock \bibinfo{journal}{\emph{arXiv preprint arXiv:1912.06680}}
  (\bibinfo{year}{2019}).
\newblock


\bibitem[\protect\citeauthoryear{Bernstein, Givan, Immerman, and
  Zilberstein}{Bernstein et~al\mbox{.}}{2002}]%
        {bernstein2002complexity}
\bibfield{author}{\bibinfo{person}{Daniel~S Bernstein}, \bibinfo{person}{Robert
  Givan}, \bibinfo{person}{Neil Immerman}, {and} \bibinfo{person}{Shlomo
  Zilberstein}.} \bibinfo{year}{2002}\natexlab{}.
\newblock \showarticletitle{The complexity of decentralized control of Markov
  decision processes}.
\newblock \bibinfo{journal}{\emph{Mathematics of operations research}}
  \bibinfo{volume}{27}, \bibinfo{number}{4} (\bibinfo{year}{2002}),
  \bibinfo{pages}{819--840}.
\newblock


\bibitem[\protect\citeauthoryear{Chen, Huang, Chiang, Chen, and Zhu}{Chen
  et~al\mbox{.}}{2022}]%
        {chen2022dgpo}
\bibfield{author}{\bibinfo{person}{Wenze Chen}, \bibinfo{person}{Shiyu Huang},
  \bibinfo{person}{Yuan Chiang}, \bibinfo{person}{Ting Chen}, {and}
  \bibinfo{person}{Jun Zhu}.} \bibinfo{year}{2022}\natexlab{}.
\newblock \showarticletitle{DGPO: Discovering Multiple Strategies with
  Diversity-Guided Policy Optimization}.
\newblock \bibinfo{journal}{\emph{arXiv preprint arXiv:2207.05631}}
  (\bibinfo{year}{2022}).
\newblock


\bibitem[\protect\citeauthoryear{Cobbe, Hesse, Hilton, and Schulman}{Cobbe
  et~al\mbox{.}}{2020}]%
        {cobbe2020leveraging}
\bibfield{author}{\bibinfo{person}{Karl Cobbe}, \bibinfo{person}{Chris Hesse},
  \bibinfo{person}{Jacob Hilton}, {and} \bibinfo{person}{John Schulman}.}
  \bibinfo{year}{2020}\natexlab{}.
\newblock \showarticletitle{Leveraging procedural generation to benchmark
  reinforcement learning}. In \bibinfo{booktitle}{\emph{International
  conference on machine learning}}. PMLR, \bibinfo{pages}{2048--2056}.
\newblock


\bibitem[\protect\citeauthoryear{Czarnecki, Gidel, Tracey, Tuyls, Omidshafiei,
  Balduzzi, and Jaderberg}{Czarnecki et~al\mbox{.}}{2020}]%
        {czarnecki2020real}
\bibfield{author}{\bibinfo{person}{Wojciech~M Czarnecki},
  \bibinfo{person}{Gauthier Gidel}, \bibinfo{person}{Brendan Tracey},
  \bibinfo{person}{Karl Tuyls}, \bibinfo{person}{Shayegan Omidshafiei},
  \bibinfo{person}{David Balduzzi}, {and} \bibinfo{person}{Max Jaderberg}.}
  \bibinfo{year}{2020}\natexlab{}.
\newblock \showarticletitle{Real world games look like spinning tops}.
\newblock \bibinfo{journal}{\emph{Advances in Neural Information Processing
  Systems}}  \bibinfo{volume}{33} (\bibinfo{year}{2020}),
  \bibinfo{pages}{17443--17454}.
\newblock


\bibitem[\protect\citeauthoryear{Ecoffet, Huizinga, Lehman, Stanley, and
  Clune}{Ecoffet et~al\mbox{.}}{2021}]%
        {ecoffet2021first}
\bibfield{author}{\bibinfo{person}{Adrien Ecoffet}, \bibinfo{person}{Joost
  Huizinga}, \bibinfo{person}{Joel Lehman}, \bibinfo{person}{Kenneth~O
  Stanley}, {and} \bibinfo{person}{Jeff Clune}.}
  \bibinfo{year}{2021}\natexlab{}.
\newblock \showarticletitle{First return, then explore}.
\newblock \bibinfo{journal}{\emph{Nature}} \bibinfo{volume}{590},
  \bibinfo{number}{7847} (\bibinfo{year}{2021}), \bibinfo{pages}{580--586}.
\newblock


\bibitem[\protect\citeauthoryear{Feng, Xie, Liu, and Wang}{Feng
  et~al\mbox{.}}{2022}]%
        {feng2022multi}
\bibfield{author}{\bibinfo{person}{Lei Feng}, \bibinfo{person}{Yuxuan Xie},
  \bibinfo{person}{Bing Liu}, {and} \bibinfo{person}{Shuyan Wang}.}
  \bibinfo{year}{2022}\natexlab{}.
\newblock \showarticletitle{Multi-Level Credit Assignment for Cooperative
  Multi-Agent Reinforcement Learning}.
\newblock \bibinfo{journal}{\emph{Applied Sciences}} \bibinfo{volume}{12},
  \bibinfo{number}{14} (\bibinfo{year}{2022}), \bibinfo{pages}{6938}.
\newblock


\bibitem[\protect\citeauthoryear{Foerster, Farquhar, Afouras, Nardelli, and
  Whiteson}{Foerster et~al\mbox{.}}{2018}]%
        {foerster2018counterfactual}
\bibfield{author}{\bibinfo{person}{Jakob Foerster}, \bibinfo{person}{Gregory
  Farquhar}, \bibinfo{person}{Triantafyllos Afouras}, \bibinfo{person}{Nantas
  Nardelli}, {and} \bibinfo{person}{Shimon Whiteson}.}
  \bibinfo{year}{2018}\natexlab{}.
\newblock \showarticletitle{Counterfactual multi-agent policy gradients}. In
  \bibinfo{booktitle}{\emph{Proceedings of the AAAI conference on artificial
  intelligence}}, Vol.~\bibinfo{volume}{32}.
\newblock


\bibitem[\protect\citeauthoryear{Fu, Yu, Xu, Yang, and Wu}{Fu
  et~al\mbox{.}}{2022}]%
        {fu2022revisiting}
\bibfield{author}{\bibinfo{person}{Wei Fu}, \bibinfo{person}{Chao Yu},
  \bibinfo{person}{Zelai Xu}, \bibinfo{person}{Jiaqi Yang}, {and}
  \bibinfo{person}{Yi Wu}.} \bibinfo{year}{2022}\natexlab{}.
\newblock \showarticletitle{Revisiting some common practices in cooperative
  multi-agent reinforcement learning}.
\newblock \bibinfo{journal}{\emph{arXiv preprint arXiv:2206.07505}}
  (\bibinfo{year}{2022}).
\newblock


\bibitem[\protect\citeauthoryear{Gao, Shi, Du, Wang, Chen, Lian, Qiu, Han,
  Wang, Ye, et~al\mbox{.}}{Gao et~al\mbox{.}}{2021}]%
        {gao2021learning}
\bibfield{author}{\bibinfo{person}{Yiming Gao}, \bibinfo{person}{Bei Shi},
  \bibinfo{person}{Xueying Du}, \bibinfo{person}{Liang Wang},
  \bibinfo{person}{Guangwei Chen}, \bibinfo{person}{Zhenjie Lian},
  \bibinfo{person}{Fuhao Qiu}, \bibinfo{person}{Guoan Han},
  \bibinfo{person}{Weixuan Wang}, \bibinfo{person}{Deheng Ye}, {et~al\mbox{.}}}
  \bibinfo{year}{2021}\natexlab{}.
\newblock \showarticletitle{Learning Diverse Policies in MOBA Games via
  Macro-Goals}.
\newblock \bibinfo{journal}{\emph{Advances in Neural Information Processing
  Systems}}  \bibinfo{volume}{34} (\bibinfo{year}{2021}),
  \bibinfo{pages}{16171--16182}.
\newblock


\bibitem[\protect\citeauthoryear{Google}{Google}{2020}]%
        {kaggle}
\bibfield{author}{\bibinfo{person}{Google}.} \bibinfo{year}{2020}\natexlab{}.
\newblock \bibinfo{title}{Google Research Football Competition 2020}.
\newblock
  \bibinfo{howpublished}{\url{https://www.kaggle.com/c/google-football}}.
\newblock


\bibitem[\protect\citeauthoryear{Guan, Liu, Hong, Zhang, Fang, Zeng, and
  Lin}{Guan et~al\mbox{.}}{2022}]%
        {guan2022perfectdou}
\bibfield{author}{\bibinfo{person}{Yang Guan}, \bibinfo{person}{Minghuan Liu},
  \bibinfo{person}{Weijun Hong}, \bibinfo{person}{Weinan Zhang},
  \bibinfo{person}{Fei Fang}, \bibinfo{person}{Guangjun Zeng}, {and}
  \bibinfo{person}{Yue Lin}.} \bibinfo{year}{2022}\natexlab{}.
\newblock \showarticletitle{PerfectDou: Dominating DouDizhu with Perfect
  Information Distillation}.
\newblock \bibinfo{journal}{\emph{arXiv preprint arXiv:2203.16406}}
  (\bibinfo{year}{2022}).
\newblock


\bibitem[\protect\citeauthoryear{Heinrich, Lanctot, and Silver}{Heinrich
  et~al\mbox{.}}{2015}]%
        {heinrich2015fictitious}
\bibfield{author}{\bibinfo{person}{Johannes Heinrich}, \bibinfo{person}{Marc
  Lanctot}, {and} \bibinfo{person}{David Silver}.}
  \bibinfo{year}{2015}\natexlab{}.
\newblock \showarticletitle{Fictitious self-play in extensive-form games}. In
  \bibinfo{booktitle}{\emph{International conference on machine learning}}.
  PMLR, \bibinfo{pages}{805--813}.
\newblock


\bibitem[\protect\citeauthoryear{Herbrich, Minka, and Graepel}{Herbrich
  et~al\mbox{.}}{2006}]%
        {herbrich2006trueskill}
\bibfield{author}{\bibinfo{person}{Ralf Herbrich}, \bibinfo{person}{Tom Minka},
  {and} \bibinfo{person}{Thore Graepel}.} \bibinfo{year}{2006}\natexlab{}.
\newblock \showarticletitle{Trueskill™: A Bayesian skill rating system}. In
  \bibinfo{booktitle}{\emph{Proceedings of the 19th international conference on
  neural information processing systems}}. \bibinfo{pages}{569--576}.
\newblock


\bibitem[\protect\citeauthoryear{Hochreiter and Schmidhuber}{Hochreiter and
  Schmidhuber}{1997}]%
        {hochreiter1997long}
\bibfield{author}{\bibinfo{person}{Sepp Hochreiter} {and}
  \bibinfo{person}{J{\"u}rgen Schmidhuber}.} \bibinfo{year}{1997}\natexlab{}.
\newblock \showarticletitle{Long short-term memory}.
\newblock \bibinfo{journal}{\emph{Neural computation}} \bibinfo{volume}{9},
  \bibinfo{number}{8} (\bibinfo{year}{1997}), \bibinfo{pages}{1735--1780}.
\newblock


\bibitem[\protect\citeauthoryear{Huang, Chen, Zhang, Li, Zhu, Ye, Chen, and
  Zhu}{Huang et~al\mbox{.}}{2021}]%
        {huang2021tikick}
\bibfield{author}{\bibinfo{person}{Shiyu Huang}, \bibinfo{person}{Wenze Chen},
  \bibinfo{person}{Longfei Zhang}, \bibinfo{person}{Ziyang Li},
  \bibinfo{person}{Fengming Zhu}, \bibinfo{person}{Deheng Ye},
  \bibinfo{person}{Ting Chen}, {and} \bibinfo{person}{Jun Zhu}.}
  \bibinfo{year}{2021}\natexlab{}.
\newblock \showarticletitle{TiKick: Towards Playing Multi-agent Football Full
  Games from Single-agent Demonstrations}.
\newblock \bibinfo{journal}{\emph{arXiv preprint arXiv:2110.04507}}
  (\bibinfo{year}{2021}).
\newblock


\bibitem[\protect\citeauthoryear{Huang, Su, Zhu, and Chen}{Huang
  et~al\mbox{.}}{2019}]%
        {huang2019combo}
\bibfield{author}{\bibinfo{person}{Shiyu Huang}, \bibinfo{person}{Hang Su},
  \bibinfo{person}{Jun Zhu}, {and} \bibinfo{person}{Ting Chen}.}
  \bibinfo{year}{2019}\natexlab{}.
\newblock \showarticletitle{Combo-Action: Training Agent For FPS Game with
  Auxiliary Tasks}. AAAI.
\newblock


\bibitem[\protect\citeauthoryear{JiDi}{JiDi}{2022}]%
        {jidi_football}
\bibfield{author}{\bibinfo{person}{JiDi}.} \bibinfo{year}{2022}\natexlab{}.
\newblock \bibinfo{title}{JiDi Olympics Football}.
\newblock
  \bibinfo{howpublished}{\url{https://github.com/jidiai/ai_lib/blob/master/env/olympics_football.py}}.
\newblock


\bibitem[\protect\citeauthoryear{Kalyanakrishnan, Liu, and
  Stone}{Kalyanakrishnan et~al\mbox{.}}{2007}]%
        {kalyanakrishnan2007half}
\bibfield{author}{\bibinfo{person}{Shivaram Kalyanakrishnan},
  \bibinfo{person}{Yaxin Liu}, {and} \bibinfo{person}{Peter Stone}.}
  \bibinfo{year}{2007}\natexlab{}.
\newblock \showarticletitle{Half field offense in RoboCup soccer: A multiagent
  reinforcement learning case study}. In \bibinfo{booktitle}{\emph{RoboCup
  2006: Robot Soccer World Cup X 10}}. Springer, \bibinfo{pages}{72--85}.
\newblock


\bibitem[\protect\citeauthoryear{Kapturowski, Campos, Jiang, Raki{\'c}evi{\'c},
  van Hasselt, Blundell, and Badia}{Kapturowski et~al\mbox{.}}{2022}]%
        {kapturowski2022human}
\bibfield{author}{\bibinfo{person}{Steven Kapturowski},
  \bibinfo{person}{V{\'\i}ctor Campos}, \bibinfo{person}{Ray Jiang},
  \bibinfo{person}{Nemanja Raki{\'c}evi{\'c}}, \bibinfo{person}{Hado van
  Hasselt}, \bibinfo{person}{Charles Blundell}, {and}
  \bibinfo{person}{Adri{\`a}~Puigdom{\`e}nech Badia}.}
  \bibinfo{year}{2022}\natexlab{}.
\newblock \showarticletitle{Human-level Atari 200x faster}.
\newblock \bibinfo{journal}{\emph{arXiv preprint arXiv:2209.07550}}
  (\bibinfo{year}{2022}).
\newblock


\bibitem[\protect\citeauthoryear{Kapturowski, Ostrovski, Quan, Munos, and
  Dabney}{Kapturowski et~al\mbox{.}}{2018}]%
        {kapturowski2018recurrent}
\bibfield{author}{\bibinfo{person}{Steven Kapturowski}, \bibinfo{person}{Georg
  Ostrovski}, \bibinfo{person}{John Quan}, \bibinfo{person}{Remi Munos}, {and}
  \bibinfo{person}{Will Dabney}.} \bibinfo{year}{2018}\natexlab{}.
\newblock \showarticletitle{Recurrent experience replay in distributed
  reinforcement learning}. In \bibinfo{booktitle}{\emph{International
  conference on learning representations}}.
\newblock


\bibitem[\protect\citeauthoryear{Kempka, Wydmuch, Runc, Toczek, and
  Ja{\'s}kowski}{Kempka et~al\mbox{.}}{2016}]%
        {kempka2016vizdoom}
\bibfield{author}{\bibinfo{person}{Micha{\l} Kempka}, \bibinfo{person}{Marek
  Wydmuch}, \bibinfo{person}{Grzegorz Runc}, \bibinfo{person}{Jakub Toczek},
  {and} \bibinfo{person}{Wojciech Ja{\'s}kowski}.}
  \bibinfo{year}{2016}\natexlab{}.
\newblock \showarticletitle{Vizdoom: A doom-based ai research platform for
  visual reinforcement learning}. In \bibinfo{booktitle}{\emph{2016 IEEE
  Conference on Computational Intelligence and Games (CIG)}}. IEEE,
  \bibinfo{pages}{1--8}.
\newblock


\bibitem[\protect\citeauthoryear{Kingma and Ba}{Kingma and Ba}{2014}]%
        {kingma2014adam}
\bibfield{author}{\bibinfo{person}{Diederik~P Kingma} {and}
  \bibinfo{person}{Jimmy Ba}.} \bibinfo{year}{2014}\natexlab{}.
\newblock \showarticletitle{Adam: A method for stochastic optimization}.
\newblock \bibinfo{journal}{\emph{arXiv preprint arXiv:1412.6980}}
  (\bibinfo{year}{2014}).
\newblock


\bibitem[\protect\citeauthoryear{Kitano, Asada, Kuniyoshi, Noda, and
  Osawa}{Kitano et~al\mbox{.}}{1997}]%
        {kitano1997robocup}
\bibfield{author}{\bibinfo{person}{Hiroaki Kitano}, \bibinfo{person}{Minoru
  Asada}, \bibinfo{person}{Yasuo Kuniyoshi}, \bibinfo{person}{Itsuki Noda},
  {and} \bibinfo{person}{Eiichi Osawa}.} \bibinfo{year}{1997}\natexlab{}.
\newblock \showarticletitle{Robocup: The robot world cup initiative}. In
  \bibinfo{booktitle}{\emph{Proceedings of the first international conference
  on Autonomous agents}}. \bibinfo{pages}{340--347}.
\newblock


\bibitem[\protect\citeauthoryear{Konda and Tsitsiklis}{Konda and
  Tsitsiklis}{1999}]%
        {konda1999actor}
\bibfield{author}{\bibinfo{person}{Vijay Konda} {and} \bibinfo{person}{John
  Tsitsiklis}.} \bibinfo{year}{1999}\natexlab{}.
\newblock \showarticletitle{Actor-critic algorithms}.
\newblock \bibinfo{journal}{\emph{Advances in neural information processing
  systems}}  \bibinfo{volume}{12} (\bibinfo{year}{1999}).
\newblock


\bibitem[\protect\citeauthoryear{Kurach, Raichuk, Stanczyk, Zajkac, Bachem,
  Espeholt, Riquelme, Vincent, Michalski, Bousquet, et~al\mbox{.}}{Kurach
  et~al\mbox{.}}{2019}]%
        {googlefootball}
\bibfield{author}{\bibinfo{person}{Karol Kurach}, \bibinfo{person}{Anton
  Raichuk}, \bibinfo{person}{Piotr Stanczyk}, \bibinfo{person}{Michal Zajkac},
  \bibinfo{person}{Olivier Bachem}, \bibinfo{person}{Lasse Espeholt},
  \bibinfo{person}{Carlos Riquelme}, \bibinfo{person}{Damien Vincent},
  \bibinfo{person}{Marcin Michalski}, \bibinfo{person}{Olivier Bousquet},
  {et~al\mbox{.}}} \bibinfo{year}{2019}\natexlab{}.
\newblock \showarticletitle{Google research football: A novel reinforcement
  learning environment}.
\newblock \bibinfo{journal}{\emph{arXiv preprint arXiv:1907.11180}}
  (\bibinfo{year}{2019}).
\newblock


\bibitem[\protect\citeauthoryear{Lanctot, Zambaldi, Gruslys, Lazaridou, Tuyls,
  P{\'e}rolat, Silver, and Graepel}{Lanctot et~al\mbox{.}}{2017}]%
        {lanctot2017unified}
\bibfield{author}{\bibinfo{person}{Marc Lanctot}, \bibinfo{person}{Vinicius
  Zambaldi}, \bibinfo{person}{Audrunas Gruslys}, \bibinfo{person}{Angeliki
  Lazaridou}, \bibinfo{person}{Karl Tuyls}, \bibinfo{person}{Julien
  P{\'e}rolat}, \bibinfo{person}{David Silver}, {and} \bibinfo{person}{Thore
  Graepel}.} \bibinfo{year}{2017}\natexlab{}.
\newblock \showarticletitle{A unified game-theoretic approach to multiagent
  reinforcement learning}.
\newblock \bibinfo{journal}{\emph{Advances in neural information processing
  systems}}  \bibinfo{volume}{30} (\bibinfo{year}{2017}).
\newblock


\bibitem[\protect\citeauthoryear{Li, Wu, Wang, Yang, Zhao, and Zhang}{Li
  et~al\mbox{.}}{2021}]%
        {li2021celebrating}
\bibfield{author}{\bibinfo{person}{Chenghao Li}, \bibinfo{person}{Chengjie Wu},
  \bibinfo{person}{Tonghan Wang}, \bibinfo{person}{Jun Yang},
  \bibinfo{person}{Qianchuan Zhao}, {and} \bibinfo{person}{Chongjie Zhang}.}
  \bibinfo{year}{2021}\natexlab{}.
\newblock \showarticletitle{Celebrating Diversity in Shared Multi-Agent
  Reinforcement Learning}.
\newblock \bibinfo{journal}{\emph{arXiv preprint arXiv:2106.02195}}
  (\bibinfo{year}{2021}).
\newblock


\bibitem[\protect\citeauthoryear{Liu, Lever, Merel, Tunyasuvunakool, Heess, and
  Graepel}{Liu et~al\mbox{.}}{2019}]%
        {liu2019emergent}
\bibfield{author}{\bibinfo{person}{Siqi Liu}, \bibinfo{person}{Guy Lever},
  \bibinfo{person}{Josh Merel}, \bibinfo{person}{Saran Tunyasuvunakool},
  \bibinfo{person}{Nicolas Heess}, {and} \bibinfo{person}{Thore Graepel}.}
  \bibinfo{year}{2019}\natexlab{}.
\newblock \showarticletitle{Emergent coordination through competition}.
\newblock \bibinfo{journal}{\emph{arXiv preprint arXiv:1902.07151}}
  (\bibinfo{year}{2019}).
\newblock


\bibitem[\protect\citeauthoryear{Liu, Lever, Wang, Merel, Eslami, Hennes,
  Czarnecki, Tassa, Omidshafiei, Abdolmaleki, et~al\mbox{.}}{Liu
  et~al\mbox{.}}{2021b}]%
        {liu2021motor}
\bibfield{author}{\bibinfo{person}{Siqi Liu}, \bibinfo{person}{Guy Lever},
  \bibinfo{person}{Zhe Wang}, \bibinfo{person}{Josh Merel}, \bibinfo{person}{SM
  Eslami}, \bibinfo{person}{Daniel Hennes}, \bibinfo{person}{Wojciech~M
  Czarnecki}, \bibinfo{person}{Yuval Tassa}, \bibinfo{person}{Shayegan
  Omidshafiei}, \bibinfo{person}{Abbas Abdolmaleki}, {et~al\mbox{.}}}
  \bibinfo{year}{2021}\natexlab{b}.
\newblock \showarticletitle{From Motor Control to Team Play in Simulated
  Humanoid Football}.
\newblock \bibinfo{journal}{\emph{arXiv preprint arXiv:2105.12196}}
  (\bibinfo{year}{2021}).
\newblock


\bibitem[\protect\citeauthoryear{Liu, Marris, Hennes, Merel, Heess, and
  Graepel}{Liu et~al\mbox{.}}{2022}]%
        {liu2022neupl}
\bibfield{author}{\bibinfo{person}{Siqi Liu}, \bibinfo{person}{Luke Marris},
  \bibinfo{person}{Daniel Hennes}, \bibinfo{person}{Josh Merel},
  \bibinfo{person}{Nicolas Heess}, {and} \bibinfo{person}{Thore Graepel}.}
  \bibinfo{year}{2022}\natexlab{}.
\newblock \showarticletitle{NeuPL: Neural Population Learning}.
\newblock \bibinfo{journal}{\emph{arXiv preprint arXiv:2202.07415}}
  (\bibinfo{year}{2022}).
\newblock


\bibitem[\protect\citeauthoryear{Liu, Jia, Wen, Yang, Hu, Chen, Fan, and
  Hu}{Liu et~al\mbox{.}}{2021a}]%
        {liu2021unifying}
\bibfield{author}{\bibinfo{person}{Xiangyu Liu}, \bibinfo{person}{Hangtian
  Jia}, \bibinfo{person}{Ying Wen}, \bibinfo{person}{Yaodong Yang},
  \bibinfo{person}{Yujing Hu}, \bibinfo{person}{Yingfeng Chen},
  \bibinfo{person}{Changjie Fan}, {and} \bibinfo{person}{Zhipeng Hu}.}
  \bibinfo{year}{2021}\natexlab{a}.
\newblock \showarticletitle{Unifying behavioral and response diversity for
  open-ended learning in zero-sum games}.
\newblock \bibinfo{journal}{\emph{arXiv preprint arXiv:2106.04958}}
  (\bibinfo{year}{2021}).
\newblock


\bibitem[\protect\citeauthoryear{Lowe, Wu, Tamar, Harb, Pieter~Abbeel, and
  Mordatch}{Lowe et~al\mbox{.}}{2017}]%
        {lowe2017multi}
\bibfield{author}{\bibinfo{person}{Ryan Lowe}, \bibinfo{person}{Yi~I Wu},
  \bibinfo{person}{Aviv Tamar}, \bibinfo{person}{Jean Harb},
  \bibinfo{person}{OpenAI Pieter~Abbeel}, {and} \bibinfo{person}{Igor
  Mordatch}.} \bibinfo{year}{2017}\natexlab{}.
\newblock \showarticletitle{Multi-agent actor-critic for mixed
  cooperative-competitive environments}.
\newblock \bibinfo{journal}{\emph{Advances in neural information processing
  systems}}  \bibinfo{volume}{30} (\bibinfo{year}{2017}).
\newblock


\bibitem[\protect\citeauthoryear{Martins, Machado, Bassani, Braga, and
  Barros}{Martins et~al\mbox{.}}{2021}]%
        {martins2021rsoccer}
\bibfield{author}{\bibinfo{person}{Felipe~B. Martins},
  \bibinfo{person}{Mateus~G. Machado}, \bibinfo{person}{Hansenclever~F.
  Bassani}, \bibinfo{person}{Pedro H.~M. Braga}, {and} \bibinfo{person}{Edna~S.
  Barros}.} \bibinfo{year}{2021}\natexlab{}.
\newblock \bibinfo{title}{rSoccer: A Framework for Studying Reinforcement
  Learning in Small and Very Small Size Robot Soccer}.
\newblock
\newblock
\showeprint[arxiv]{2106.12895}~[cs.LG]


\bibitem[\protect\citeauthoryear{Mnih, Kavukcuoglu, Silver, Rusu, Veness,
  Bellemare, Graves, Riedmiller, Fidjeland, Ostrovski, et~al\mbox{.}}{Mnih
  et~al\mbox{.}}{2015}]%
        {mnih2015human}
\bibfield{author}{\bibinfo{person}{Volodymyr Mnih}, \bibinfo{person}{Koray
  Kavukcuoglu}, \bibinfo{person}{David Silver}, \bibinfo{person}{Andrei~A
  Rusu}, \bibinfo{person}{Joel Veness}, \bibinfo{person}{Marc~G Bellemare},
  \bibinfo{person}{Alex Graves}, \bibinfo{person}{Martin Riedmiller},
  \bibinfo{person}{Andreas~K Fidjeland}, \bibinfo{person}{Georg Ostrovski},
  {et~al\mbox{.}}} \bibinfo{year}{2015}\natexlab{}.
\newblock \showarticletitle{Human-level control through deep reinforcement
  learning}.
\newblock \bibinfo{journal}{\emph{Nature}} \bibinfo{volume}{518},
  \bibinfo{number}{7540} (\bibinfo{year}{2015}), \bibinfo{pages}{529}.
\newblock


\bibitem[\protect\citeauthoryear{Ng, Harada, and Russell}{Ng
  et~al\mbox{.}}{1999}]%
        {ng1999policy}
\bibfield{author}{\bibinfo{person}{Andrew~Y Ng}, \bibinfo{person}{Daishi
  Harada}, {and} \bibinfo{person}{Stuart Russell}.}
  \bibinfo{year}{1999}\natexlab{}.
\newblock \showarticletitle{Policy invariance under reward transformations:
  Theory and application to reward shaping}. In
  \bibinfo{booktitle}{\emph{Icml}}, Vol.~\bibinfo{volume}{99}.
  \bibinfo{pages}{278--287}.
\newblock


\bibitem[\protect\citeauthoryear{NVIDIA}{NVIDIA}{2016}]%
        {nccl}
\bibfield{author}{\bibinfo{person}{NVIDIA}.} \bibinfo{year}{2016}\natexlab{}.
\newblock \bibinfo{title}{NCCL}.
\newblock \bibinfo{howpublished}{\url{https://github.com/NVIDIA/nccl}}.
\newblock


\bibitem[\protect\citeauthoryear{Parker-Holder, Pacchiano, Choromanski, and
  Roberts}{Parker-Holder et~al\mbox{.}}{2020}]%
        {parker2020effective}
\bibfield{author}{\bibinfo{person}{Jack Parker-Holder}, \bibinfo{person}{Aldo
  Pacchiano}, \bibinfo{person}{Krzysztof~M Choromanski}, {and}
  \bibinfo{person}{Stephen~J Roberts}.} \bibinfo{year}{2020}\natexlab{}.
\newblock \showarticletitle{Effective diversity in population based
  reinforcement learning}.
\newblock \bibinfo{journal}{\emph{Advances in Neural Information Processing
  Systems}}  \bibinfo{volume}{33} (\bibinfo{year}{2020}),
  \bibinfo{pages}{18050--18062}.
\newblock


\bibitem[\protect\citeauthoryear{Paszke, Gross, Massa, Lerer, Bradbury, Chanan,
  Killeen, Lin, Gimelshein, Antiga, et~al\mbox{.}}{Paszke
  et~al\mbox{.}}{2019}]%
        {paszke2019pytorch}
\bibfield{author}{\bibinfo{person}{Adam Paszke}, \bibinfo{person}{Sam Gross},
  \bibinfo{person}{Francisco Massa}, \bibinfo{person}{Adam Lerer},
  \bibinfo{person}{James Bradbury}, \bibinfo{person}{Gregory Chanan},
  \bibinfo{person}{Trevor Killeen}, \bibinfo{person}{Zeming Lin},
  \bibinfo{person}{Natalia Gimelshein}, \bibinfo{person}{Luca Antiga},
  {et~al\mbox{.}}} \bibinfo{year}{2019}\natexlab{}.
\newblock \showarticletitle{Pytorch: An imperative style, high-performance deep
  learning library}.
\newblock \bibinfo{journal}{\emph{Advances in neural information processing
  systems}}  \bibinfo{volume}{32} (\bibinfo{year}{2019}),
  \bibinfo{pages}{8026--8037}.
\newblock


\bibitem[\protect\citeauthoryear{Pearce and Zhu}{Pearce and Zhu}{2022}]%
        {CSGO2022}
\bibfield{author}{\bibinfo{person}{Tim Pearce} {and} \bibinfo{person}{Jun
  Zhu}.} \bibinfo{year}{2022}\natexlab{}.
\newblock \showarticletitle{Counter-Strike Deathmatch with Large-Scale
  Behavioural Cloning}.
\newblock \bibinfo{journal}{\emph{2022 IEEE Conference on Games (CoG)}}
  (\bibinfo{year}{2022}), \bibinfo{pages}{104--111}.
\newblock
\urldef\tempurl%
\url{https://doi.org/10.1109/CoG51982.2022.9893617}
\showDOI{\tempurl}


\bibitem[\protect\citeauthoryear{Rashid, Samvelyan, Schroeder, Farquhar,
  Foerster, and Whiteson}{Rashid et~al\mbox{.}}{2018}]%
        {rashid2018qmix}
\bibfield{author}{\bibinfo{person}{Tabish Rashid}, \bibinfo{person}{Mikayel
  Samvelyan}, \bibinfo{person}{Christian Schroeder}, \bibinfo{person}{Gregory
  Farquhar}, \bibinfo{person}{Jakob Foerster}, {and} \bibinfo{person}{Shimon
  Whiteson}.} \bibinfo{year}{2018}\natexlab{}.
\newblock \showarticletitle{Qmix: Monotonic value function factorisation for
  deep multi-agent reinforcement learning}. In
  \bibinfo{booktitle}{\emph{International Conference on Machine Learning}}.
  PMLR, \bibinfo{pages}{4295--4304}.
\newblock


\bibitem[\protect\citeauthoryear{Saxe, McClelland, and Ganguli}{Saxe
  et~al\mbox{.}}{2013}]%
        {saxe2013exact}
\bibfield{author}{\bibinfo{person}{Andrew~M Saxe}, \bibinfo{person}{James~L
  McClelland}, {and} \bibinfo{person}{Surya Ganguli}.}
  \bibinfo{year}{2013}\natexlab{}.
\newblock \showarticletitle{Exact solutions to the nonlinear dynamics of
  learning in deep linear neural networks}.
\newblock \bibinfo{journal}{\emph{arXiv preprint arXiv:1312.6120}}
  (\bibinfo{year}{2013}).
\newblock


\bibitem[\protect\citeauthoryear{Schulman, Moritz, Levine, Jordan, and
  Abbeel}{Schulman et~al\mbox{.}}{2015}]%
        {schulman2015high}
\bibfield{author}{\bibinfo{person}{John Schulman}, \bibinfo{person}{Philipp
  Moritz}, \bibinfo{person}{Sergey Levine}, \bibinfo{person}{Michael Jordan},
  {and} \bibinfo{person}{Pieter Abbeel}.} \bibinfo{year}{2015}\natexlab{}.
\newblock \showarticletitle{High-dimensional continuous control using
  generalized advantage estimation}.
\newblock \bibinfo{journal}{\emph{arXiv preprint arXiv:1506.02438}}
  (\bibinfo{year}{2015}).
\newblock


\bibitem[\protect\citeauthoryear{Schulman, Wolski, Dhariwal, Radford, and
  Klimov}{Schulman et~al\mbox{.}}{2017}]%
        {schulman2017proximal}
\bibfield{author}{\bibinfo{person}{John Schulman}, \bibinfo{person}{Filip
  Wolski}, \bibinfo{person}{Prafulla Dhariwal}, \bibinfo{person}{Alec Radford},
  {and} \bibinfo{person}{Oleg Klimov}.} \bibinfo{year}{2017}\natexlab{}.
\newblock \showarticletitle{Proximal policy optimization algorithms}.
\newblock \bibinfo{journal}{\emph{arXiv preprint arXiv:1707.06347}}
  (\bibinfo{year}{2017}).
\newblock


\bibitem[\protect\citeauthoryear{Seccia, Foglino, Leonetti, and
  Sagratella}{Seccia et~al\mbox{.}}{2022}]%
        {seccia2022novel}
\bibfield{author}{\bibinfo{person}{Ruggiero Seccia}, \bibinfo{person}{Francesco
  Foglino}, \bibinfo{person}{Matteo Leonetti}, {and} \bibinfo{person}{Simone
  Sagratella}.} \bibinfo{year}{2022}\natexlab{}.
\newblock \showarticletitle{A novel optimization perspective to the problem of
  designing sequences of tasks in a reinforcement learning framework}.
\newblock \bibinfo{journal}{\emph{Optimization and Engineering}}
  (\bibinfo{year}{2022}), \bibinfo{pages}{1--16}.
\newblock


\bibitem[\protect\citeauthoryear{Silver, Huang, Maddison, Guez, Sifre, Van
  Den~Driessche, Schrittwieser, Antonoglou, Panneershelvam, Lanctot,
  et~al\mbox{.}}{Silver et~al\mbox{.}}{2016}]%
        {silver2016mastering}
\bibfield{author}{\bibinfo{person}{David Silver}, \bibinfo{person}{Aja Huang},
  \bibinfo{person}{Chris~J Maddison}, \bibinfo{person}{Arthur Guez},
  \bibinfo{person}{Laurent Sifre}, \bibinfo{person}{George Van Den~Driessche},
  \bibinfo{person}{Julian Schrittwieser}, \bibinfo{person}{Ioannis Antonoglou},
  \bibinfo{person}{Veda Panneershelvam}, \bibinfo{person}{Marc Lanctot},
  {et~al\mbox{.}}} \bibinfo{year}{2016}\natexlab{}.
\newblock \showarticletitle{Mastering the game of Go with deep neural networks
  and tree search}.
\newblock \bibinfo{journal}{\emph{Nature}} \bibinfo{volume}{529},
  \bibinfo{number}{7587} (\bibinfo{year}{2016}), \bibinfo{pages}{484--489}.
\newblock


\bibitem[\protect\citeauthoryear{Silver, Hubert, Schrittwieser, Antonoglou,
  Lai, Guez, Lanctot, Sifre, Kumaran, Graepel, et~al\mbox{.}}{Silver
  et~al\mbox{.}}{2017}]%
        {silver2017mastering}
\bibfield{author}{\bibinfo{person}{David Silver}, \bibinfo{person}{Thomas
  Hubert}, \bibinfo{person}{Julian Schrittwieser}, \bibinfo{person}{Ioannis
  Antonoglou}, \bibinfo{person}{Matthew Lai}, \bibinfo{person}{Arthur Guez},
  \bibinfo{person}{Marc Lanctot}, \bibinfo{person}{Laurent Sifre},
  \bibinfo{person}{Dharshan Kumaran}, \bibinfo{person}{Thore Graepel},
  {et~al\mbox{.}}} \bibinfo{year}{2017}\natexlab{}.
\newblock \showarticletitle{Mastering chess and shogi by self-play with a
  general reinforcement learning algorithm}.
\newblock \bibinfo{journal}{\emph{arXiv preprint arXiv:1712.01815}}
  (\bibinfo{year}{2017}).
\newblock


\bibitem[\protect\citeauthoryear{Silver, Hubert, Schrittwieser, Antonoglou,
  Lai, Guez, Lanctot, Sifre, Kumaran, Graepel, et~al\mbox{.}}{Silver
  et~al\mbox{.}}{2018}]%
        {silver2018general}
\bibfield{author}{\bibinfo{person}{David Silver}, \bibinfo{person}{Thomas
  Hubert}, \bibinfo{person}{Julian Schrittwieser}, \bibinfo{person}{Ioannis
  Antonoglou}, \bibinfo{person}{Matthew Lai}, \bibinfo{person}{Arthur Guez},
  \bibinfo{person}{Marc Lanctot}, \bibinfo{person}{Laurent Sifre},
  \bibinfo{person}{Dharshan Kumaran}, \bibinfo{person}{Thore Graepel},
  {et~al\mbox{.}}} \bibinfo{year}{2018}\natexlab{}.
\newblock \showarticletitle{A general reinforcement learning algorithm that
  masters chess, shogi, and Go through self-play}.
\newblock \bibinfo{journal}{\emph{Science}} \bibinfo{volume}{362},
  \bibinfo{number}{6419} (\bibinfo{year}{2018}), \bibinfo{pages}{1140--1144}.
\newblock


\bibitem[\protect\citeauthoryear{Son, Kim, Kang, Hostallero, and Yi}{Son
  et~al\mbox{.}}{2019}]%
        {son2019qtran}
\bibfield{author}{\bibinfo{person}{Kyunghwan Son}, \bibinfo{person}{Daewoo
  Kim}, \bibinfo{person}{Wan~Ju Kang}, \bibinfo{person}{David~Earl Hostallero},
  {and} \bibinfo{person}{Yung Yi}.} \bibinfo{year}{2019}\natexlab{}.
\newblock \showarticletitle{Qtran: Learning to factorize with transformation
  for cooperative multi-agent reinforcement learning}. In
  \bibinfo{booktitle}{\emph{International Conference on Machine Learning}}.
  PMLR, \bibinfo{pages}{5887--5896}.
\newblock


\bibitem[\protect\citeauthoryear{Stone, Sutton, and Kuhlmann}{Stone
  et~al\mbox{.}}{2005}]%
        {stone2005reinforcement}
\bibfield{author}{\bibinfo{person}{Peter Stone}, \bibinfo{person}{Richard~S
  Sutton}, {and} \bibinfo{person}{Gregory Kuhlmann}.}
  \bibinfo{year}{2005}\natexlab{}.
\newblock \showarticletitle{Reinforcement learning for robocup soccer
  keepaway}.
\newblock \bibinfo{journal}{\emph{Adaptive Behavior}} \bibinfo{volume}{13},
  \bibinfo{number}{3} (\bibinfo{year}{2005}), \bibinfo{pages}{165--188}.
\newblock


\bibitem[\protect\citeauthoryear{Sun, Devlin, Beck, Hofmann, and Whiteson}{Sun
  et~al\mbox{.}}{2022}]%
        {sun2022JMAPPO}
\bibfield{author}{\bibinfo{person}{Mingfei Sun}, \bibinfo{person}{Sam Devlin},
  \bibinfo{person}{Jacob Beck}, \bibinfo{person}{Katja Hofmann}, {and}
  \bibinfo{person}{Shimon Whiteson}.} \bibinfo{year}{2022}\natexlab{}.
\newblock \showarticletitle{Monotonic Improvement Guarantees under
  Non-stationarity for Decentralized PPO}.
\newblock \bibinfo{journal}{\emph{arXiv preprint arXiv:2202.00082}}
  (\bibinfo{year}{2022}).
\newblock


\bibitem[\protect\citeauthoryear{Sunehag, Lever, Gruslys, Czarnecki, Zambaldi,
  Jaderberg, Lanctot, Sonnerat, Leibo, Tuyls, et~al\mbox{.}}{Sunehag
  et~al\mbox{.}}{2017}]%
        {sunehag2017value}
\bibfield{author}{\bibinfo{person}{Peter Sunehag}, \bibinfo{person}{Guy Lever},
  \bibinfo{person}{Audrunas Gruslys}, \bibinfo{person}{Wojciech~Marian
  Czarnecki}, \bibinfo{person}{Vinicius Zambaldi}, \bibinfo{person}{Max
  Jaderberg}, \bibinfo{person}{Marc Lanctot}, \bibinfo{person}{Nicolas
  Sonnerat}, \bibinfo{person}{Joel~Z Leibo}, \bibinfo{person}{Karl Tuyls},
  {et~al\mbox{.}}} \bibinfo{year}{2017}\natexlab{}.
\newblock \showarticletitle{Value-decomposition networks for cooperative
  multi-agent learning}.
\newblock \bibinfo{journal}{\emph{arXiv preprint arXiv:1706.05296}}
  (\bibinfo{year}{2017}).
\newblock


\bibitem[\protect\citeauthoryear{Taiga, Fedus, Machado, Courville, and
  Bellemare}{Taiga et~al\mbox{.}}{2019}]%
        {taiga2019bonus}
\bibfield{author}{\bibinfo{person}{Adrien~Ali Taiga}, \bibinfo{person}{William
  Fedus}, \bibinfo{person}{Marlos~C Machado}, \bibinfo{person}{Aaron
  Courville}, {and} \bibinfo{person}{Marc~G Bellemare}.}
  \bibinfo{year}{2019}\natexlab{}.
\newblock \showarticletitle{On bonus based exploration methods in the arcade
  learning environment}. In \bibinfo{booktitle}{\emph{International Conference
  on Learning Representations}}.
\newblock


\bibitem[\protect\citeauthoryear{Vinyals, Babuschkin, Czarnecki, Mathieu,
  Dudzik, Chung, Choi, Powell, Ewalds, Georgiev, et~al\mbox{.}}{Vinyals
  et~al\mbox{.}}{2019}]%
        {vinyals2019grandmaster}
\bibfield{author}{\bibinfo{person}{Oriol Vinyals}, \bibinfo{person}{Igor
  Babuschkin}, \bibinfo{person}{Wojciech~M Czarnecki},
  \bibinfo{person}{Micha{\"e}l Mathieu}, \bibinfo{person}{Andrew Dudzik},
  \bibinfo{person}{Junyoung Chung}, \bibinfo{person}{David~H Choi},
  \bibinfo{person}{Richard Powell}, \bibinfo{person}{Timo Ewalds},
  \bibinfo{person}{Petko Georgiev}, {et~al\mbox{.}}}
  \bibinfo{year}{2019}\natexlab{}.
\newblock \showarticletitle{Grandmaster level in StarCraft II using multi-agent
  reinforcement learning}.
\newblock \bibinfo{journal}{\emph{Nature}} \bibinfo{volume}{575},
  \bibinfo{number}{7782} (\bibinfo{year}{2019}), \bibinfo{pages}{350--354}.
\newblock


\bibitem[\protect\citeauthoryear{w9PcJLyb}{w9PcJLyb}{2020}]%
        {kaggle_15}
\bibfield{author}{\bibinfo{person}{w9PcJLyb}.} \bibinfo{year}{2020}\natexlab{}.
\newblock \bibinfo{title}{Solution ranked 15th in Kaggle Football Competition}.
\newblock \bibinfo{howpublished}{\url{https://github.com/w9PcJLyb/GFootball}}.
\newblock


\bibitem[\protect\citeauthoryear{Wang, Ren, Liu, Yu, and Zhang}{Wang
  et~al\mbox{.}}{2020b}]%
        {wang2020qplex}
\bibfield{author}{\bibinfo{person}{Jianhao Wang}, \bibinfo{person}{Zhizhou
  Ren}, \bibinfo{person}{Terry Liu}, \bibinfo{person}{Yang Yu}, {and}
  \bibinfo{person}{Chongjie Zhang}.} \bibinfo{year}{2020}\natexlab{b}.
\newblock \showarticletitle{Qplex: Duplex dueling multi-agent q-learning}.
\newblock \bibinfo{journal}{\emph{arXiv preprint arXiv:2008.01062}}
  (\bibinfo{year}{2020}).
\newblock


\bibitem[\protect\citeauthoryear{Wang, Zhang, Hu, Wang, Zhang, Gao, Hao, Lv,
  and Fan}{Wang et~al\mbox{.}}{2022}]%
        {wang2022individual}
\bibfield{author}{\bibinfo{person}{Li Wang}, \bibinfo{person}{Yupeng Zhang},
  \bibinfo{person}{Yujing Hu}, \bibinfo{person}{Weixun Wang},
  \bibinfo{person}{Chongjie Zhang}, \bibinfo{person}{Yang Gao},
  \bibinfo{person}{Jianye Hao}, \bibinfo{person}{Tangjie Lv}, {and}
  \bibinfo{person}{Changjie Fan}.} \bibinfo{year}{2022}\natexlab{}.
\newblock \showarticletitle{Individual Reward Assisted Multi-Agent
  Reinforcement Learning}. In \bibinfo{booktitle}{\emph{International
  Conference on Machine Learning}}. PMLR, \bibinfo{pages}{23417--23432}.
\newblock


\bibitem[\protect\citeauthoryear{Wang, Wu, Evans, Tenenbaum, Parkes, and
  Kleiman-Weiner}{Wang et~al\mbox{.}}{2020c}]%
        {wang2020too}
\bibfield{author}{\bibinfo{person}{Rose~E Wang}, \bibinfo{person}{Sarah~A Wu},
  \bibinfo{person}{James~A Evans}, \bibinfo{person}{Joshua~B Tenenbaum},
  \bibinfo{person}{David~C Parkes}, {and} \bibinfo{person}{Max
  Kleiman-Weiner}.} \bibinfo{year}{2020}\natexlab{c}.
\newblock \showarticletitle{Too many cooks: Coordinating multi-agent
  collaboration through inverse planning}.
\newblock  (\bibinfo{year}{2020}).
\newblock


\bibitem[\protect\citeauthoryear{Wang, Gupta, Mahajan, Peng, Whiteson, and
  Zhang}{Wang et~al\mbox{.}}{2020a}]%
        {wang2020rode}
\bibfield{author}{\bibinfo{person}{Tonghan Wang}, \bibinfo{person}{Tarun
  Gupta}, \bibinfo{person}{Anuj Mahajan}, \bibinfo{person}{Bei Peng},
  \bibinfo{person}{Shimon Whiteson}, {and} \bibinfo{person}{Chongjie Zhang}.}
  \bibinfo{year}{2020}\natexlab{a}.
\newblock \showarticletitle{RODE: Learning Roles to Decompose Multi-Agent
  Tasks}.
\newblock \bibinfo{journal}{\emph{arXiv preprint arXiv:2010.01523}}
  (\bibinfo{year}{2020}).
\newblock


\bibitem[\protect\citeauthoryear{Wen, Kuba, Lin, Zhang, Wen, Wang, and
  Yang}{Wen et~al\mbox{.}}{2022}]%
        {wen2022multi}
\bibfield{author}{\bibinfo{person}{Muning Wen}, \bibinfo{person}{Jakub~Grudzien
  Kuba}, \bibinfo{person}{Runji Lin}, \bibinfo{person}{Weinan Zhang},
  \bibinfo{person}{Ying Wen}, \bibinfo{person}{Jun Wang}, {and}
  \bibinfo{person}{Yaodong Yang}.} \bibinfo{year}{2022}\natexlab{}.
\newblock \showarticletitle{Multi-Agent Reinforcement Learning is a Sequence
  Modeling Problem}.
\newblock \bibinfo{journal}{\emph{arXiv preprint arXiv:2205.14953}}
  (\bibinfo{year}{2022}).
\newblock


\bibitem[\protect\citeauthoryear{Wu and Tian}{Wu and Tian}{2016}]%
        {wu2016training}
\bibfield{author}{\bibinfo{person}{Yuxin Wu} {and} \bibinfo{person}{Yuandong
  Tian}.} \bibinfo{year}{2016}\natexlab{}.
\newblock \showarticletitle{Training agent for first-person shooter game with
  actor-critic curriculum learning}.
\newblock  (\bibinfo{year}{2016}).
\newblock


\bibitem[\protect\citeauthoryear{Ye, Chen, Zhang, Chen, Yuan, Liu, Chen, Liu,
  Qiu, Yu, et~al\mbox{.}}{Ye et~al\mbox{.}}{2020}]%
        {ye2020towards}
\bibfield{author}{\bibinfo{person}{Deheng Ye}, \bibinfo{person}{Guibin Chen},
  \bibinfo{person}{Wen Zhang}, \bibinfo{person}{Sheng Chen},
  \bibinfo{person}{Bo Yuan}, \bibinfo{person}{Bo Liu}, \bibinfo{person}{Jia
  Chen}, \bibinfo{person}{Zhao Liu}, \bibinfo{person}{Fuhao Qiu},
  \bibinfo{person}{Hongsheng Yu}, {et~al\mbox{.}}}
  \bibinfo{year}{2020}\natexlab{}.
\newblock \showarticletitle{Towards playing full moba games with deep
  reinforcement learning}.
\newblock \bibinfo{journal}{\emph{Advances in Neural Information Processing
  Systems}}  \bibinfo{volume}{33} (\bibinfo{year}{2020}),
  \bibinfo{pages}{621--632}.
\newblock


\bibitem[\protect\citeauthoryear{Yu, Velu, Vinitsky, Wang, Bayen, and Wu}{Yu
  et~al\mbox{.}}{2021}]%
        {yu2021surprising}
\bibfield{author}{\bibinfo{person}{Chao Yu}, \bibinfo{person}{Akash Velu},
  \bibinfo{person}{Eugene Vinitsky}, \bibinfo{person}{Yu Wang},
  \bibinfo{person}{Alexandre Bayen}, {and} \bibinfo{person}{Yi Wu}.}
  \bibinfo{year}{2021}\natexlab{}.
\newblock \showarticletitle{The Surprising Effectiveness of MAPPO in
  Cooperative, Multi-Agent Games}.
\newblock \bibinfo{journal}{\emph{arXiv preprint arXiv:2103.01955}}
  (\bibinfo{year}{2021}).
\newblock


\bibitem[\protect\citeauthoryear{Zha, Xie, Ma, Zhang, Lian, Hu, and Liu}{Zha
  et~al\mbox{.}}{2021}]%
        {zha2021douzero}
\bibfield{author}{\bibinfo{person}{Daochen Zha}, \bibinfo{person}{Jingru Xie},
  \bibinfo{person}{Wenye Ma}, \bibinfo{person}{Sheng Zhang},
  \bibinfo{person}{Xiangru Lian}, \bibinfo{person}{Xia Hu}, {and}
  \bibinfo{person}{Ji Liu}.} \bibinfo{year}{2021}\natexlab{}.
\newblock \showarticletitle{Douzero: Mastering doudizhu with self-play deep
  reinforcement learning}. In \bibinfo{booktitle}{\emph{International
  Conference on Machine Learning}}. PMLR, \bibinfo{pages}{12333--12344}.
\newblock


\bibitem[\protect\citeauthoryear{Zhang, Xu, Wang, Wu, Keutzer, Gonzalez, and
  Tian}{Zhang et~al\mbox{.}}{2020}]%
        {zhang2020bebold}
\bibfield{author}{\bibinfo{person}{Tianjun Zhang}, \bibinfo{person}{Huazhe Xu},
  \bibinfo{person}{Xiaolong Wang}, \bibinfo{person}{Yi Wu},
  \bibinfo{person}{Kurt Keutzer}, \bibinfo{person}{Joseph~E Gonzalez}, {and}
  \bibinfo{person}{Yuandong Tian}.} \bibinfo{year}{2020}\natexlab{}.
\newblock \showarticletitle{BeBold: Exploration Beyond the Boundary of Explored
  Regions}.
\newblock \bibinfo{journal}{\emph{arXiv preprint arXiv:2012.08621}}
  (\bibinfo{year}{2020}).
\newblock


\bibitem[\protect\citeauthoryear{Zheng, Li, Mao, and Tei}{Zheng
  et~al\mbox{.}}{2022}]%
        {zheng2022local}
\bibfield{author}{\bibinfo{person}{Nianzhao Zheng}, \bibinfo{person}{Jialong
  Li}, \bibinfo{person}{Zhenyu Mao}, {and} \bibinfo{person}{Kenji Tei}.}
  \bibinfo{year}{2022}\natexlab{}.
\newblock \showarticletitle{From Local to Global: A Curriculum Learning
  Approach for Reinforcement Learning-based Traffic Signal Control}. In
  \bibinfo{booktitle}{\emph{2022 IEEE 2nd International Conference on Software
  Engineering and Artificial Intelligence (SEAI)}}. IEEE,
  \bibinfo{pages}{253--258}.
\newblock


\bibitem[\protect\citeauthoryear{Zhou, Liu, Sui, Li, and Chung}{Zhou
  et~al\mbox{.}}{2020}]%
        {zhou2020learning}
\bibfield{author}{\bibinfo{person}{Meng Zhou}, \bibinfo{person}{Ziyu Liu},
  \bibinfo{person}{Pengwei Sui}, \bibinfo{person}{Yixuan Li}, {and}
  \bibinfo{person}{Yuk~Ying Chung}.} \bibinfo{year}{2020}\natexlab{}.
\newblock \showarticletitle{Learning implicit credit assignment for cooperative
  multi-agent reinforcement learning}.
\newblock \bibinfo{journal}{\emph{Advances in Neural Information Processing
  Systems}}  \bibinfo{volume}{33} (\bibinfo{year}{2020}),
  \bibinfo{pages}{11853--11864}.
\newblock


\bibitem[\protect\citeauthoryear{Ziyang~Li}{Ziyang~Li}{2020}]%
        {wekick}
\bibfield{author}{\bibinfo{person}{Fengming~Zhu Ziyang~Li, Kaiwen~Zhu}.}
  \bibinfo{year}{2020}\natexlab{}.
\newblock \bibinfo{title}{WeKick}.
\newblock
  \bibinfo{howpublished}{\url{https://www.kaggle.com/c/google-football/discussion/202232}}.
\newblock


\end{thebibliography}
\clearpage

\includepdf[pages=-]{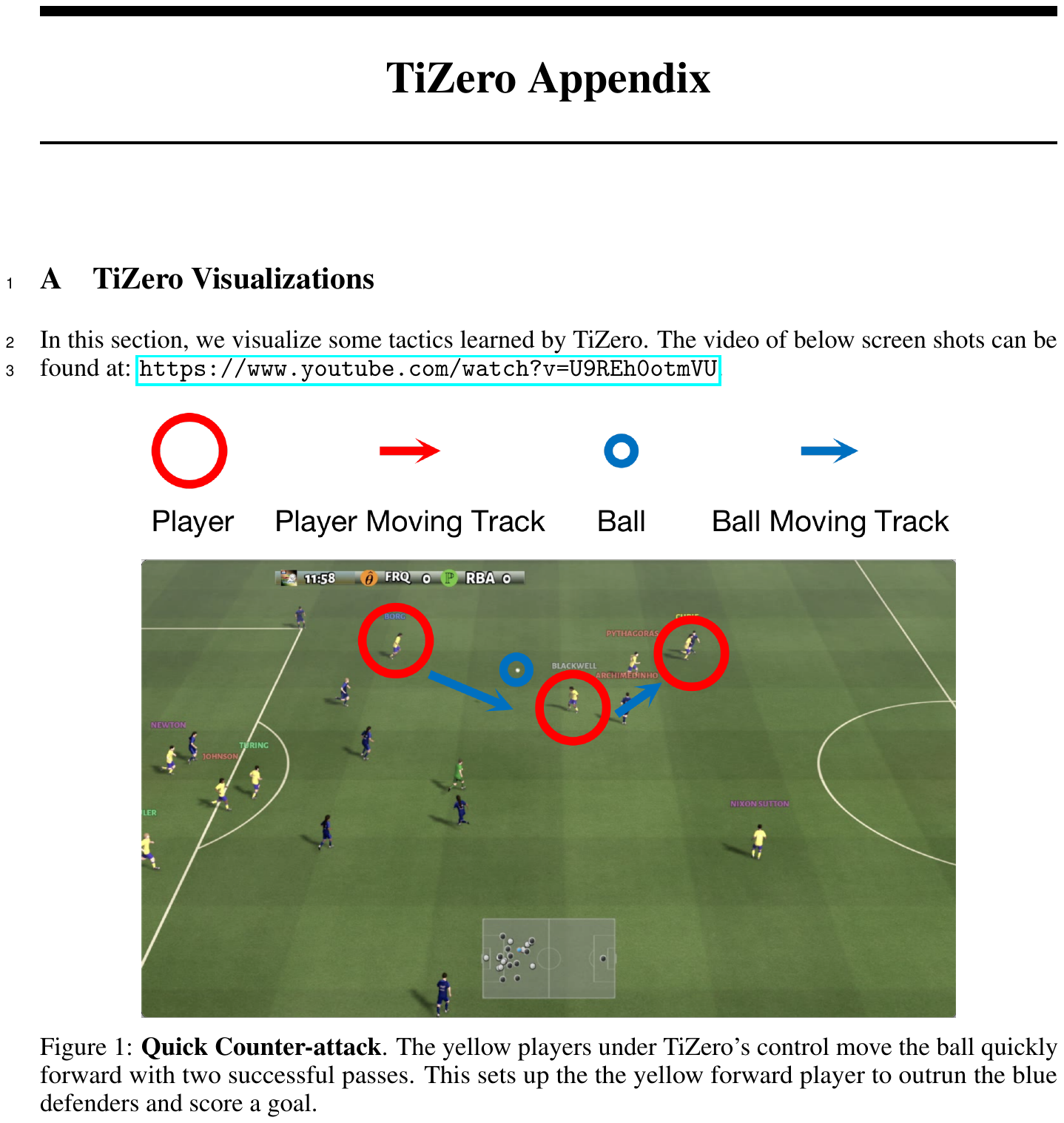}

\end{document}